\documentclass{article}

\usepackage{microtype}
\usepackage{graphicx}
\usepackage{subfigure}
\usepackage{booktabs} 

\usepackage{hyperref}

\usepackage[accepted]{icml2025}

\usepackage{url}            
\usepackage{amsfonts}       
\usepackage{nicefrac}       
\usepackage{xcolor}         
\usepackage{color, colortbl}
\usepackage{tcolorbox}
\usepackage{xspace}
\usepackage{bm} 
\usepackage{multirow}
\usepackage{array}
\usepackage{amsfonts}
\usepackage{listings}
\usepackage{threeparttable}
\usepackage{wrapfig}
\usepackage{algorithm}
\usepackage{diagbox}
\usepackage{pifont}
\usepackage{tcolorbox}

\usepackage{amsmath}
\usepackage{amssymb}
\usepackage{mathtools}
\usepackage{amsthm}
\usepackage{tabularx}
\newcommand{\newl}{\textcolor{gray}{\textbackslash n }}
\newcommand{\promptfield}[1]{\textcolor{blue}{$\langle$#1$\rangle$}}

\definecolor{cyberpink}{RGB}{235,110, 152}
\definecolor{lightgray}{rgb}{0.93,0.93,0.93}

\newcommand{\qrag}[0]{\textsc{QueryRAG}\xspace}
\newcommand{\lrag}[0]{\textsc{LabelRAG}\xspace}
\newcommand{\frag}[0]{\textsc{FewshotRAG}\xspace}

\newcommand{\R}[1]{{%
    \textbf{%
        \ifstrequal{#1}{1}{\textcolor{red}{R#1}}{%
        \ifstrequal{#1}{2}{\textcolor{blue}{R#1}}{%
        \ifstrequal{#1}{3}{\textcolor{magenta}{R#1}}{%
        \ifstrequal{#1}{4}{\textcolor{teal}{R#1}}{%
                           \textcolor{cyan}{R#1}%
        }}}}%
    }%
}}
\definecolor{BlueGreen}{rgb}{0.0, 0.87, 0.87}
\definecolor{WildStrawberry}{rgb}{1.0, 0.26, 0.64}
\definecolor{lightblue}{RGB}{220, 235, 255}  
\definecolor{lightpink}{RGB}{255, 225, 235}  

\newtcbox{\hlprimarytab}{on line, box align=base, colback=lightblue, size=fbox, before upper=\strut, top=-1.5pt, bottom=-1.5pt, left=-2pt, right=-2pt, boxrule=0pt}
\newtcbox{\hlsecondarytab}{on line, box align=base, colback=lightpink,size=fbox, before upper=\strut, top=-1.5pt, bottom=-1.5pt, left=-2pt, right=-2pt, boxrule=0pt}
\newtcbox{\hlthirdarytab}{on line, box align=base, colback=cyberpink!80,size=fbox, before upper=\strut, top=-1.5pt, bottom=-1.5pt, left=-2pt, right=-2pt, boxrule=0pt}
\newcommand{\orange}[1]{{\hlsecondarytab{#1}}}
\newcommand{\blue}[1]{{\hlprimarytab{#1}}}
\newcommand{\pink}[1]{{\hlthirdarytab{#1}}}

\definecolor{gold}{HTML}{FFD700}
\definecolor{brown}{HTML}{C0C0C0}

\newcommand{\first}[1]{{\textbf{\textcolor{red!60} {#1}}}} 
\newcommand{\second}[1]{{\textbf{\textcolor{blue!60} {#1}}}}  
\newcommand{\third}[1]{\underline{\textbf{\textcolor{brown} {#1}}}}

\usepackage[capitalize,noabbrev]{cleveref}

\theoremstyle{plain}

\theoremstyle{definition}

\theoremstyle{remark}

\icmltitlerunning{Submission and Formatting Instructions for ICML 2025}

\begin{document}

\twocolumn[
  \icmltitle{Are Large Language Models In-Context Graph Learners?}



  \icmlsetsymbol{equal}{*}

  \begin{icmlauthorlist}
    \icmlauthor{Jintang Li}{yyy}
    \icmlauthor{Ruofan Wu}{comp}
    \icmlauthor{Yuchang Zhu}{yyy}
    \icmlauthor{Huizhe Zhang}{yyy}
    \icmlauthor{Liang Chen}{yyy}
    \icmlauthor{Zibin Zheng}{yyy}
  \end{icmlauthorlist}

  \icmlaffiliation{yyy}{Sun Yat-sen University}
  \icmlaffiliation{comp}{Coupang}
  \icmlcorrespondingauthor{Jintang Li}{lijt55@mail2.sysu.edu.cn}

  \icmlkeywords{Machine Learning, ICML}

  \vskip 0.3in
]



\printAffiliationsAndNotice{}  

\begin{abstract}
  Large language models (LLMs) have demonstrated remarkable in-context reasoning capabilities across a wide range of tasks, particularly with unstructured inputs such as language or images. However, LLMs struggle to handle structured data, such as graphs, due to their lack of understanding of non-Euclidean structures. As a result, without additional fine-tuning, their performance significantly lags behind that of graph neural networks (GNNs) in graph learning tasks. In this paper, we show that learning on graph data can be conceptualized as a retrieval-augmented generation (RAG) process, where specific instances (e.g., nodes or edges) act as queries, and the graph itself serves as the retrieved context. Building on this insight, we propose a series of RAG frameworks to enhance the in-context learning capabilities of LLMs for graph learning tasks. Comprehensive evaluations demonstrate that our proposed RAG frameworks significantly improve LLM performance on graph-based tasks, particularly in scenarios where a pretrained LLM must be used without modification or accessed via an API.

\end{abstract}

\section{Introduction}

Large language models (LLMs) exhibit astonishing reasoning capabilities across a wide variety of real-world tasks, including text generation~\cite{qwen2.5}, code completion~\cite{deepseek-coder}, and question answering~\cite{rag}. The widespread success of LLMs is often attributed to their in-context learning capabilities~\cite{icl_survey}. Building on top of just raw reasoning abilities, LLMs can solve a new task for which they have not been explicitly trained, by being provided with a few examples (few-shot)~\cite{few_shot} or even with instructions describing the task (zero-shot)~\cite{zero_shot_cot}.

Leveraging their success in handling unstructured data, LLMs have increasingly been applied to tasks involving structured data, such as graphs~\cite{ChenMLJWWWYFLT23,instructglm}.
Essentially, language models are trained on semantic text as next-token predictors, which are inherently capable of learning from any textual input, including graph-like structures represented as text. For example, graphs can be encoded into textual representations, such as edge lists, adjacency matrices, or even graph descriptions in natural language~\cite{instructglm,GPT4Graph}. When presented with such textual representations, LLMs are expected to understand the relationships and properties of the graph, extending their reasoning abilities to the graph domain.

However, breaking down the graph representation into verbal semantic prompts (i.e., graph tokenization) leads to the loss of critical structural information~\cite{GPT4Graph}. We argue that the intrinsic relationships and dependencies between nodes and edges require a deeper understanding of spatial and relational structures, which cannot be fully captured through textual encoding alone. As a result, despite their impressive performance on unstructured tasks, LLMs without additional fine-tuning struggle to match the effectiveness of specialized models like graph neural networks (GNNs)~\cite{gcn,gat}, which are explicitly designed to process graph data and leverage its structural properties.

\begin{figure}[t]
  \centering
  \includegraphics[width=\linewidth]{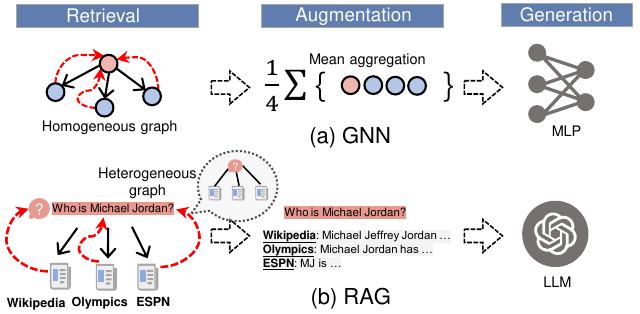}
  \vspace{-4mm}
  \caption{A technical comparison between \blue{GNN} and \orange{RAG}. Both leverage contextual information beyond the raw input through\textit{homogeneous} and \textit{heterogeneous} structures, respectively.}
  \label{fig:comparison}
\end{figure}

As LLMs continue to evolve, new techniques such as retrieval-augmented generation (RAG)\cite{rag} and chain-of-thought (CoT) reasoning\cite{cot} have been introduced to mitigate their limitations in reasoning. These techniques demonstrate that LLMs can be strong learners by leveraging context-dependent semantics or rationales.
For graph data, message-passing GNNs (hereafter referred to as GNNs for brevity) have emerged as one of the most effective tools. They propagate and aggregate information from contextual inputs~\cite{message_passing} in an inductive manner, a process conceptually similar to in-context learning~\cite{graphsage}. While LLMs excel at understanding context and reasoning with unstructured data, their in-context learning capabilities for graph-structured data remain largely unexplored. This naturally raises the question: \textit{are LLMs effective in-context graph learners?}

In this work, we investigate the in-context learning capabilities of LLMs on graph data. Initial results indicate that LLMs are not effective as in-context graph learners. Even state-of-the-art LLMs fail to match the performance of GNNs in node classification tasks. To understand this gap, we analyze the key principles that make GNNs effective for graph data, particularly their ability to capture local structural information and propagate it across the graph.
Our findings reveal that \textit{message passing, the core mechanism of GNNs, can be interpreted as a recursive RAG step applied to each node along with its graph context} (see Figure~\ref{fig:comparison} for an illustration). This RAG-like process enables even a simple neural network, such as an MLP, to achieve strong performance on graph-related tasks by leveraging retrieved contextual information effectively.

Building on this insight, we propose a series of RAG-based frameworks—\qrag, \lrag, and \frag—that leverage graph structure as inherent context to enhance the in-context learning capabilities of LLMs. Unlike standard RAG approaches that require external retrieval mechanisms, our frameworks inherently retrieve relevant context by utilizing the local graph neighborhood of each node. \qrag retrieves only node features from graph neighbors, while \lrag incorporates both node features and corresponding labels. \frag extends this further by combining query node information with retrieved contextual information to provide richer in-context learning signals.

We conduct extensive experiments on multiple text-attributed graph datasets to evaluate the effectiveness of our frameworks. Results show that LLMs augmented with our RAG frameworks significantly outperform their zero-shot and standard few-shot counterparts. Notably, \lrag and \frag can even match or exceed the performance of supervised MLPs and, in some cases, approach the effectiveness of GNNs or fine-tuned graph LLMs. These findings suggest that retrieval-augmented in-context learning is a viable alternative to traditional supervised learning, especially in scenarios where fine-tuning is not feasible.

\textbf{Contributions.} We outline our main contributions below.
\begin{itemize}
  \item \textbf{GNNs are RAG networks.} We are the first to relate message-passing GNNs to RAG, both of which are powerful tools for facilitating context-aware tasks. By drawing this connection, we provide new insights into leveraging contextual information for LLMs from proximal queries and corresponding labels.
  \item \textbf{Graph-guided RAG frameworks.} Motivated by the connection between RAG and message-passing GNNs, we introduce three specialized RAG frameworks: \qrag, \lrag, and \frag. These frameworks leverage the query, label, or both from retrieved nodes to enhance the understanding of LLMs on graph data.
  \item \textbf{Experimental results.}
        We conduct extensive experiments with LLMs on node classification tasks to evaluate the effectiveness of our proposed RAG frameworks. Experimental results show that, with our frameworks, an off-the-shelf LLM can match or even surpass the performance of supervised GNNs and fine-tuned graph LLMs.
\end{itemize}
Our work highlights that, although language models are not explicitly designed for graph learning tasks, their generalization ability, few-shot learning capabilities, and understanding of relational structures in text enable them to perform graph-related tasks with minimal yet informative examples or instructions. This makes them powerful tools for both language and graph-based reasoning, bridging the gap between textual and graph-structured data.

\begin{figure*}
  \centering
  \includegraphics[width=\linewidth]{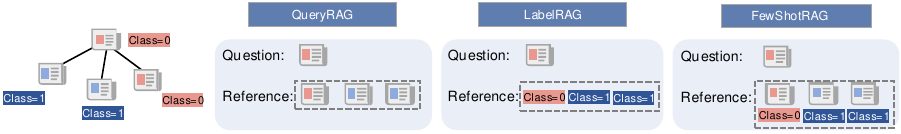}
  \vspace{-5mm}
  \caption{Illustration of the proposed \qrag, \lrag, and \frag frameworks. Specifically, \qrag uses the retrieved queries as references, \lrag incorporates only the corresponding labels, and \frag combines both query and label information as few-shot context..}
  \label{fig:framework}
\end{figure*}

\section{Preliminary}
In this section, we introduce the notation and formalize key concepts related to (large) language models and graph neural networks for learning over text-attributed graphs.

\paragraph{Notations.}
We define a text-attributed graph $\mathcal{G}=(\mathcal{V}, \mathcal{E}, \mathcal{T})$ with N nodes, where $\mathcal{V}$ is the set of nodes and $\mathcal{E} \subseteq (\mathcal{V} \times \mathcal{V})$ is set of edges connecting them. For each node $v_i \in \mathcal{V}$ is associated with a textual attribute $x_i$, and $\mathcal{X} = \{x_i|v_i\in \mathcal{V}\}$ is the attribute set.
We consider a fundamental task, namely node classification, which predicts the class label of a node $v_i$. This task can be conceptualized and approached as a text classification problem but with additional graph context. For simplicity, let $\mathcal{Y} = {y_i}_{i=1}^N$ denote the set of ground-truth labels, where $y_i \in \mathcal{C}$ for node classification, with $\mathcal{C}$ representing the set of class labels.

\paragraph{In-context learning.}
In-context learning is a paradigm that enables language models to learn tasks using only a few examples as demonstrations. Formally, given a pretrained LLM, the model is expected to generate an output  $r$  for a new query input  $q$  by conditioning on a prompt that includes in-context examples along with the query input, i.e.,
\begin{equation}
  r = \operatorname{LLM}(\mathcal{C}_K, q),
\end{equation}
where $\mathcal{C}_K=(q_1, r_1, . . . , q_K, r_K)$ is the demonstration set and $K$ is the number of demonstrations. Each pair of input-output examples $(q_i,r_i) \in \mathcal{C}_K$, also called shots, is sampled from a known distribution or a training dataset. Without loss of generality, we omit the task-specific instructions for the generation task hereafter for simplicity.
Depending on whether the $K$-shot demonstration examples belong to the same task as the query, in-context learning (ICL) can be categorized as task-specific ICL or cross-task ICL~\cite{icl_survey2}. This capability is quite intriguing as it allows models to adapt to a wide range of downstream tasks on-the-fly—i.e., without the need to perform any parameter updates after the model is trained.

\paragraph{Retrieval-augmented generation.}
RAG combines a retrieval mechanism with a generative model (e.g., a LLM) to generate responses by leveraging external knowledge. Formally, given a query $q$ and a corpus of documents $\mathcal{D}$, the RAG process can be formulated as:
\begin{equation}
  r = \operatorname{LLM}(q, \{c_i\}_{i=1}^K), \  \text{s.t.} \  D(q, c_i)<\epsilon,
\end{equation}
where $\{c_i\}_{i=1}^K$ represents the set of top-$K$ documents or chunks retrieved from the corpus $\mathcal{D}$. We use the distance metric $D(\cdot)$ and a sufficiently small constant $\epsilon$ to represent the retrieval step, which computes the relevance of the textual corpus in $\mathcal{D}$ to the query $q$ and selects the top-$K$ documents or chunks to serve as additional context for LLMs. The distance function $D(\cdot)$ can be implemented as cosine similarity, i.e., $D(x,y)=\frac{x\cdot y}{\|x\|\|y\|}$.

\paragraph{Graph neural networks.}
Most GNNs rely on iteratively propagating information between neighboring nodes in the graph. This process follows the message-passing paradigm~\cite{message_passing}, where nodes exchange vector-based messages along edges and update their representations by aggregating the received messages in a permutation-invariant manner. In general, the message-passing step in GNNs is formalized as follows:
\begin{align}
  \label{eq:gnn_mp}
  h_i               =\operatorname{UPD}\left(\left\{x_i, \operatorname{AGG}\left(\left\{x_j: v_j \in \mathcal{N}(v_i)\right\}\right)\right\}\right),
\end{align}
where $\mathcal{N}(v_i)$ denotes the neighborhood set of node $v_i$.
$\operatorname{AGG}$ and $\operatorname{UPD}$ are the aggregation and update function, respectively. Basically, both $\operatorname{AGG}$ and $\operatorname{UPD}$ can be learnable functions or neural networks, which leads to different variants of GNNs~\cite{gat,sgc,graphsage}.
We misuse $x_j$ here as the vector attributes of node $v_j$, which can be encoded from its corresponding textual attributes by language models. Since an MLP network can be considered a special case of a GNN, we can formulate the MLP below based on Eq.~\eqref{eq:gnn_mp} as \( h_i = \operatorname{UPD}(x_i) \).

\section{Present work}
In this section, we begin with an empirical study to investigate the in-context learning capabilities of LLMs in graph learning tasks. Next, we establish the connection between RAG and GNNs. Finally, we introduce three novel RAG frameworks designed to improve LLMs’ understanding of graph data, as illustrated in Figure~\ref{fig:framework}.

\subsection{Empirical study}
We introduce a concrete node classification task on graphs and investigate prompting techniques in in-context learning settings. Our goal is to explore the potential of LLMs to reason on graph data without any additional fine-tuning, focusing solely on their pure in-context capabilities.

\paragraph{Experimental settings.}
We experiment with two representative text-attributed graph datasets: Cora and Pubmed~\cite{sen2008collective}. For the node classification task, we use \textsc{Llama-3.1-8B-Instruct}\cite{llama3} and \textsc{DeepSeek-V3}\cite{deepseekv3} as base models, representing state-of-the-art open-source and closed-source large language models (LLMs) for natural language tasks, respectively. Our approach to in-context learning follows \cite{ChenMLJWWWYFLT23,few_shot}; however, in this work, we systematically explore additional settings to enhance learning within the context.
\begin{itemize}
  \item Few-shot: The model is provided with a few ( $K$ ) demonstrations of the task during inference as conditioning~\cite{few_shot}, but no weight updates are performed. In addition to the target node’s content, this approach incorporates the content and labels of \textit{randomly} selected in-context examples from the training set. In our experiments, we set  $K=3$.
  \item One-shot: Similar to few-shot learning, but with $K=1$, meaning only a single demonstration is provided.
  \item Zero-shot: Similar to few-shot learning but without any in-context examples. Instead, only a natural language description of the task is provided.
  \item RAG: We include RAG as an in-context baseline, where the most similar  $K$  samples are retrieved to augment the prompt. Since no external corpus is available for retrieval, we instead retrieve the textual content of similar nodes from the graph. This differs from traditional RAG, which retrieves \textit{in-context references} from an \textit{external} knowledge source.
\end{itemize}
For comparison, we include supervised GNNs (i.e., GCN) and MLP as baselines. Both models are trained in a supervised manner on the training set, using 20 samples per class~\cite{gcn,tsgfm}. Due to space limitations, we defer the detailed experimental settings to the Appendix~\ref{sec:exp_setting}.

\begin{figure}
  \centering
  \subfigure[Cora]{\includegraphics[width=0.45\linewidth]{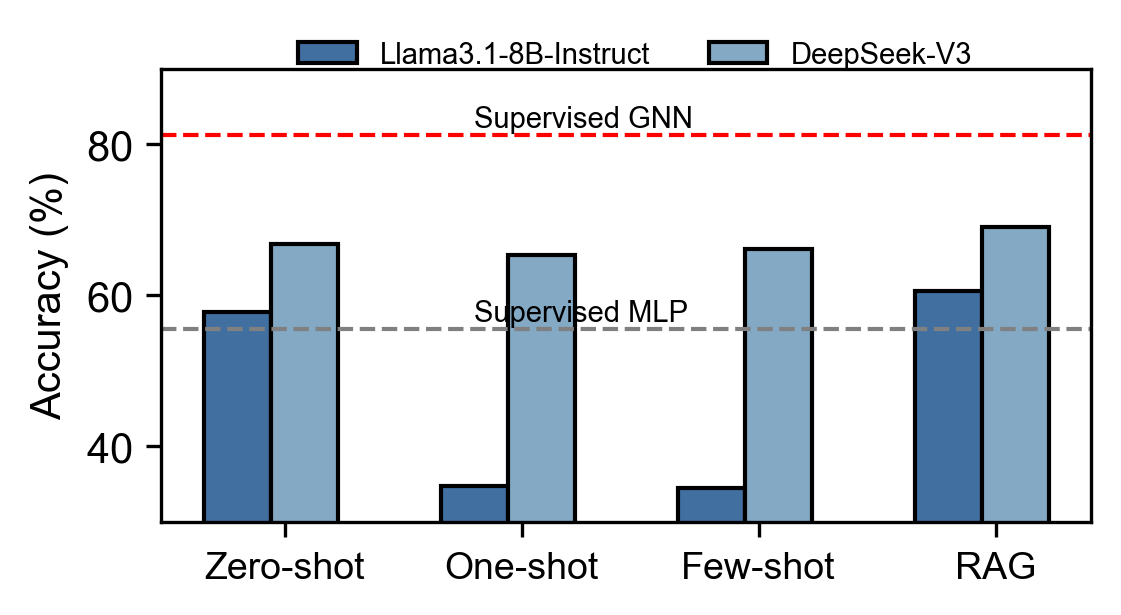}}
  \subfigure[Pubmed]{\includegraphics[width=0.45\linewidth]{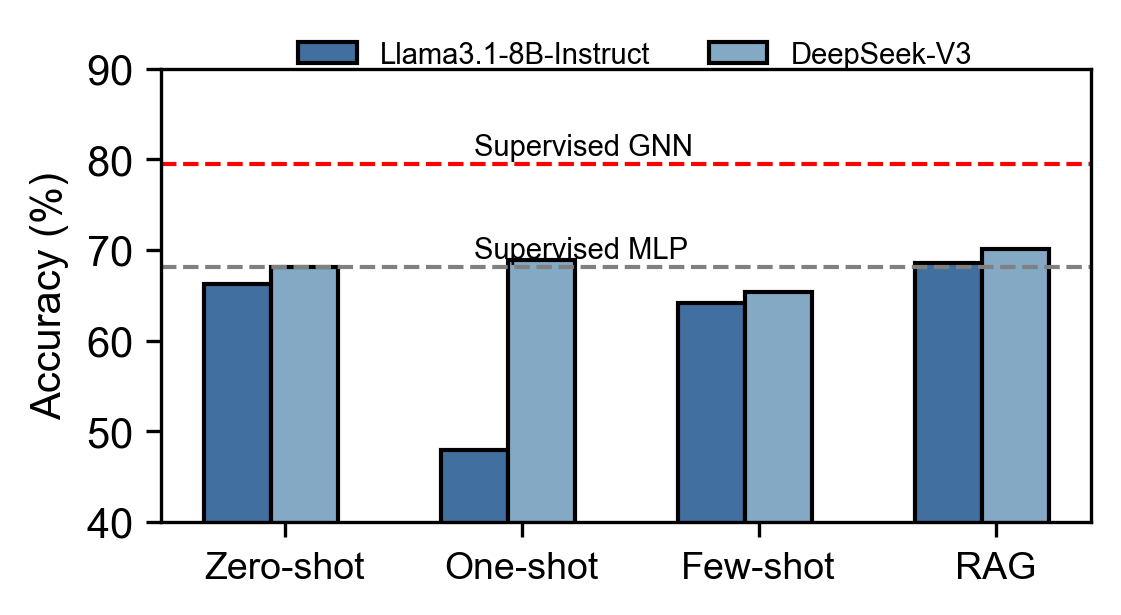}}
  \caption{In-context node classification results on Cora and Pubmed datasets. GNN and MLP are supervisedly trained on partial nodes.}
  \label{fig:empirical_baseline}
\end{figure}

\begin{figure}
  \centering
  \subfigure[Cora]{\includegraphics[width=0.45\linewidth]{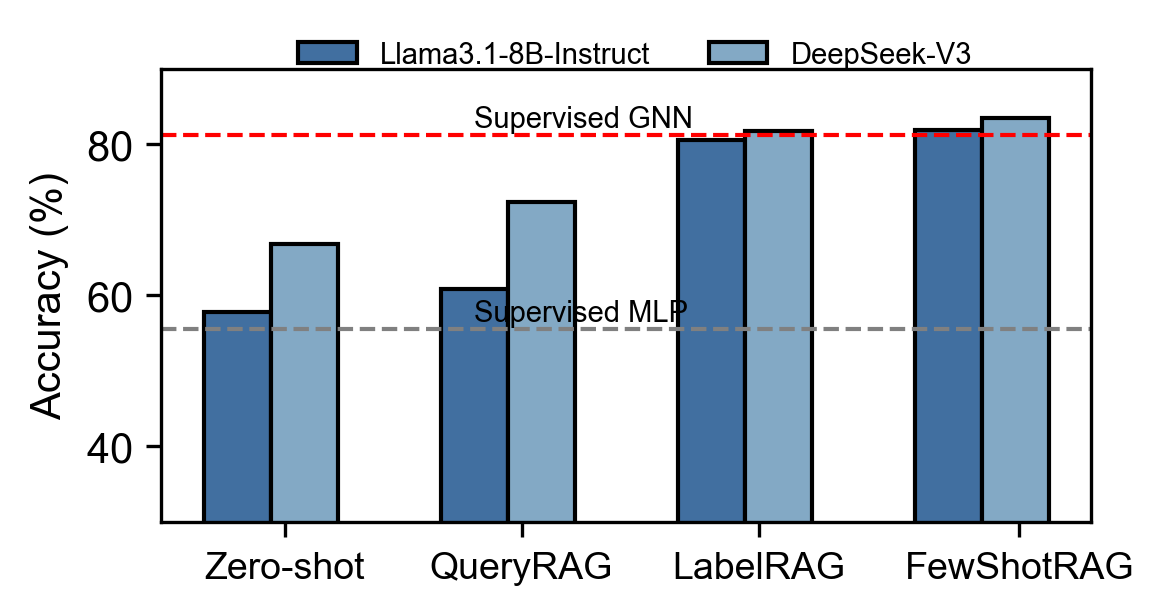}}
  \subfigure[Pubmed]{\includegraphics[width=0.45\linewidth]{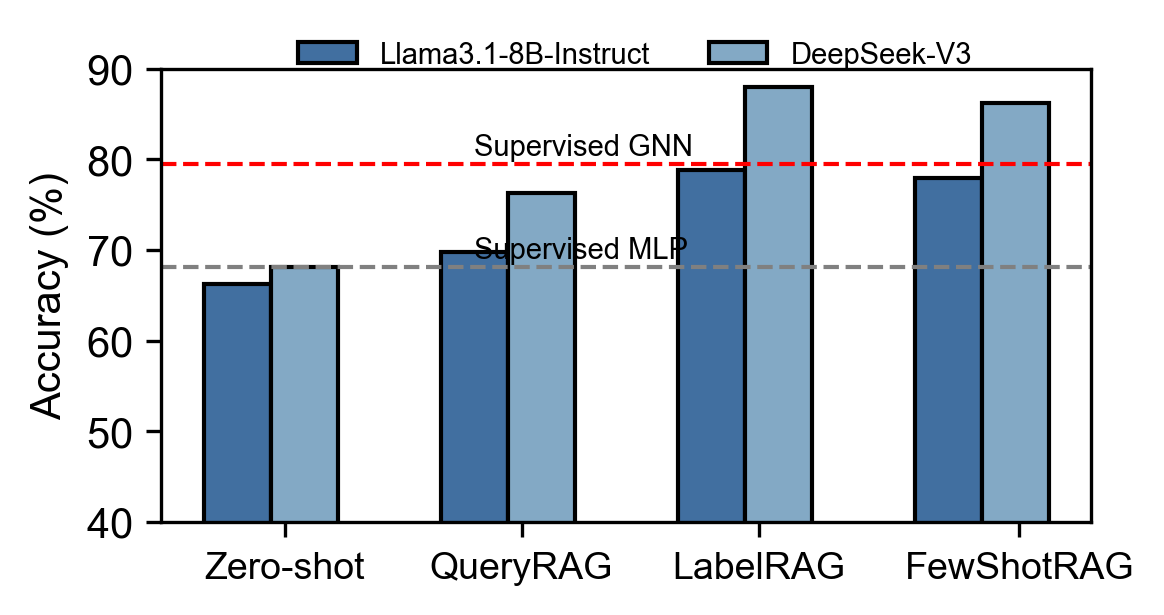}}
  \caption{In-context node classification results on Cora and Pubmed datasets with our proposed \qrag, \lrag, and \frag.}
  \label{fig:empirical}
\end{figure}

\paragraph{Observation I: LLMs are not inherently effective in-context graph learners.}
The in-context learning results of \textsc{Llama-3.1-8B-Instruct} and \textsc{DeepSeek-V3} on text-attributed graphs are presented in Figure~\ref{fig:empirical}. Both LLMs perform significantly worse than supervised baselines such as GNNs in the zero-shot setting, highlighting the challenge of graph-based reasoning without explicit in-context examples. More surprisingly, in the one-shot and few-shot settings, both LLMs exhibit a further performance drop compared to zero-shot, particularly on the Cora dataset. This suggests that LLMs may struggle to effectively utilize in-context examples, and in some cases, these examples can even negatively impact their performance.

\paragraph{Observation II: RAG enhances LLM performance on graph learning.}
Compared to the zero-shot and few-shot settings, RAG achieves higher accuracy for both LLMs, indicating that incorporating graph node-specific textual content as context benefits LLM-based reasoning. LLMs augmented with RAG show improved performance and even surpass the supervised MLP on both datasets. However, RAG still falls significantly behind supervised GNNs, which highlights the limitations of in-context learning for graph tasks compared to models explicitly trained on node features. This underscores the gap between in-context learning and fully supervised methods in graph learning, suggesting that while retrieval-based augmentation helps, it does not yet fully bridge the performance disparity.

Overall, the findings suggest that LLMs have limited but promising in-context learning capabilities for the node classification task. This highlights the need for further research to improve their reasoning abilities on graph-structured data without requiring additional fine-tuning.

\subsection{Connecting GNNs with RAG}
In our empirical experiments, RAG generally achieves the best performance among in-context baselines, although it still significantly underperforms supervised GNNs. This motivates us to explore potential improvements in RAG for graph learning tasks.

In the literature, it has been shown that improving the quality of retrieved context significantly enhances the performance of LLMs~\cite{DBLP:journals/tacl/RamLDMSLS23}. However, in RAG, the textual attributes of nodes (i.e., the \textit{queries}) are treated as an external \textit{corpus} to retrieve relevant textual context for each node, which differs from traditional RAG methods~\cite{rag}. In this regard, the objective of retrieving `neighborhood' information from the graph has led us to explore the connections between RAG and GNNs.

Both RAG and GNNs are context-based methods that leverage contextual information in a similar way. Specifically, RAG retrieves relevant information from an external corpus based on the query to serve as context, while GNNs aggregate information from the local neighborhood of the query node within a graph. Given these similarities, we are motivated to establish a formal connection between RAG and GNNs for further exploration. To this end, we first rewrite the formulations of RAG and GNNs below.
\begin{align}
  \label{eq:rqg2}
  r_i & = \orange{$\operatorname{GEN}$}\left(\left\{q_i, \orange{$\operatorname{AUG}$}\left(\left\{c_j: c_j \in \orange{$\mathcal{D}(q_i)$}\right\}\right)\right\}\right), \\
  \label{eq:gnn2}
  h_i & =\blue{$\operatorname{UPD}$}\left(\left\{x_i, \blue{$\operatorname{AGG}$}\left(\left\{x_j: v_j \in \blue{$\mathcal{N}(v_i)$}\right\}\right)\right\}\right),
\end{align}
where {$\operatorname{GEN}$} and {$\operatorname{AUG}$} are the generation and augmentation step in RAG. Typically, $\operatorname{GEN}$ is implemented with an off-the-shelf LLM while $\operatorname{AUG}$ involves prompt engineering to condition retrieved contexts.  $\mathcal{D}(q_i)=\{q_j | D(q_i, q_j)< \epsilon\}$ is the set of retrieved documents that are close to the query $q_i$. From Eqs.~\eqref{eq:rqg2} and \eqref{eq:gnn2}, we observe that RAG and GNNs share a similar formulation across the \textit{retrieval}, \textit{augmentation}, and \textit{generation} steps.

\subsection{Graph-guided RAG frameworks}
Building on the inherent connections between GNNs and RAG, we propose three graph-guided RAG frameworks: \qrag, \lrag, and \frag. Specifically, \qrag utilizes the \textit{query} context of neighboring nodes from the graph, while \lrag incorporates the \textit{label} information. Combining these two approaches, \frag leverages both \textit{query} and \textit{label} information from graph-guided contexts.

\paragraph{From GNN to \qrag.}
Although we have demonstrated that GNNs can be conceptualized as RAG networks, there are key differences between them. Unlike RAG, where context is dynamically retrieved from an external corpus, the context in GNNs is inherently relational, relying on the graph’s topology and interactions between nodes. This makes GNNs particularly well-suited for graph learning tasks, where relationships and dependencies play a critical role.
Furthermore, the retrieved context in GNNs is more akin to retrieving a \textit{similar query} based on node features or representations, rather than relying on an external corpus, as is typical in vanilla RAG. In this regard, we propose the concept of \qrag, which utilizes the queries from other queries as the external corpus for retrieval:\begin{align}
  \label{eq:query_rag}
  r_i & = \operatorname{GEN}\left(\left\{q_i, \operatorname{AUG}\left(\left\{\pink{$q_j$}: v_j \in \mathcal{N}(v_i)\right\}\right)\right\}\right).
\end{align}
In \qrag, we take inspiration from GNNs and utilize the text attributes of neighboring nodes $\mathcal{N}$  as inherently retrieved context. Each neighboring node contributes to the contextual understanding of the query node. By treating the local graph neighborhood as a source of context, \qrag combines the strengths of GNN-style aggregation with the flexibility of generative models. This design enables \qrag to effectively handle tasks that require relational reasoning.
Additionally, this inherently relational retrieval mechanism simplifies the workflow, as it does not depend on an external corpus or complex similarity-based retrieval pipelines. In \qrag,  $q_i$  can represent a specific query in natural language understanding tasks or the textual attributes of a node in node classification tasks.

\paragraph{From label propagation to \lrag.}
Since we have established connections between RAG and GNNs, particularly message-passing GNNs, we can conceptualize graph learning as an RAG-specific task and leverage different message-passing architectures to develop advanced RAG variants. For instance, \qrag utilizes only the `query' information from neighboring or retrieved nodes, leaving label information unexplored.
In fact, propagating label information among adjacent nodes is an effective message-passing strategy. This process, commonly known as label propagation, iteratively updates node labels by aggregating information from their neighbors~\cite{lp}. Formally, label propagation at each iteration can be expressed in a message-passing framework:
\begin{equation}
  \begin{aligned}
    \hat{y}_i & =  (1 - \alpha) y_i + \alpha \sum_{v_j \in \mathcal{N}(v_i)}  y_j,
    \\
              & =\operatorname{UPD}\left(\left\{y_i, \operatorname{AGG}\left(\left\{y_j: v_j \in \mathcal{N}(v_i)\right\}\right)\right\}\right),
  \end{aligned}
\end{equation}
where $\alpha$ is a hyperparameter balancing the influence of neighboring labels and the initial label.
Recent studies have also shown that propagating node labels during message passing in GNNs can be an effective strategy~\cite{ShiHFZWS21,c_and_s,hetero2net}. Building on this insight, we propose \lrag, which extends the concept of label propagation to RAG learning:
\begin{align}
  \label{eq:label_rag}
  r_i & = \operatorname{GEN}\left(\left\{q_i, \operatorname{AUG}\left(\left\{\pink{$y_j\ \text{or}\ r_j$}: v_j \in \mathcal{N}(v_i)\right\}\right)\right\}\right),
\end{align}
where $y_i\ \text{or}\ r_j$ can be the ground-truth label or response of query $q_i$.
In \lrag, we incorporate label information as part of the retrieved context to enhance LLMs’ comprehension of the query. This mechanism aligns with the principles of label propagation in GNNs, where labels from neighboring nodes contribute to a node’s representation. However, \lrag extends this idea to retrieval-augmented frameworks by embedding label context directly into the prompts.
The augmented context, $\operatorname{AUG}(\{y_j: v_j \in \mathcal{N}(v_i)\})$, provides semantically meaningful information that enriches the generative model’s understanding of the query  $q_i$. This enables the model to leverage both the semantic information from the query and the relational label context derived from the graph structure.
This approach is particularly beneficial in scenarios where label correlations and dependencies play a crucial role, as it allows the model to capture structured relationships beyond the query’s standalone textual features.

\paragraph{\frag: few-shot learning on graphs.}
Both \qrag and \lrag utilize external query and label information as contextual inputs to enhance the generation process, respectively. They follow similar principles to few-shot in-context learning, where  $K$  examples, consisting of queries and their corresponding responses, are provided to improve LLM comprehension and performance.
Since incorporating query and label information as external context helps bridge the gap between the input and the desired output, a natural extension is to combine \qrag and \lrag in a manner similar to few-shot learning:
\begin{align}
  \label{eq:few_shot_rag}
  r_i & = \operatorname{GEN}\left(\left\{q_i, \operatorname{AUG}\left(\left\{\pink{$(q_j, y_j)$}: v_j \in \mathcal{N}(v_i)\right\}\right)\right\}\right),
\end{align}
where the retrieved tuples $(q_j, y_j)$ are used as context for the query $q_i$. Eq.~\eqref{eq:few_shot_rag} illustrates the concept of leveraging both query and label information from neighboring nodes as in-context examples, which we refer to as \frag. \frag enhances LLM reasoning in graph-based tasks by providing richer contextual guidance through both semantic query information and relational label context.

\paragraph{Remarks.}
\qrag, \lrag, and \frag are RAG architectures that integrate insights from in-context learning and message-passing GNNs in different ways. We provide a comparison of various in-context learning and message-passing approaches in Table~\ref{tab:comparison}. A given query  $q_i$  may represent the textual attributes of a node in a node classification task or a specific question in a question-answering task. Similarly,  $r_i$  can correspond to the ground-truth label of a node or the generated response.
We use the one-hop neighbors of a node as the retrieved context for the query; however, the context can be extended to include multi-hop neighbors or defined based on proximity-based retrieval methods~\cite{rag}. For comparison, we present the results on Cora and Pubmed in Figure~\ref{fig:comparison}. The promising results suggest that incorporating structured graph context provides valuable guidance for LLMs in reasoning over graph data.

\begin{figure*}
  \centering
  \subfigure[Fitness]{\includegraphics[width=0.16\linewidth]{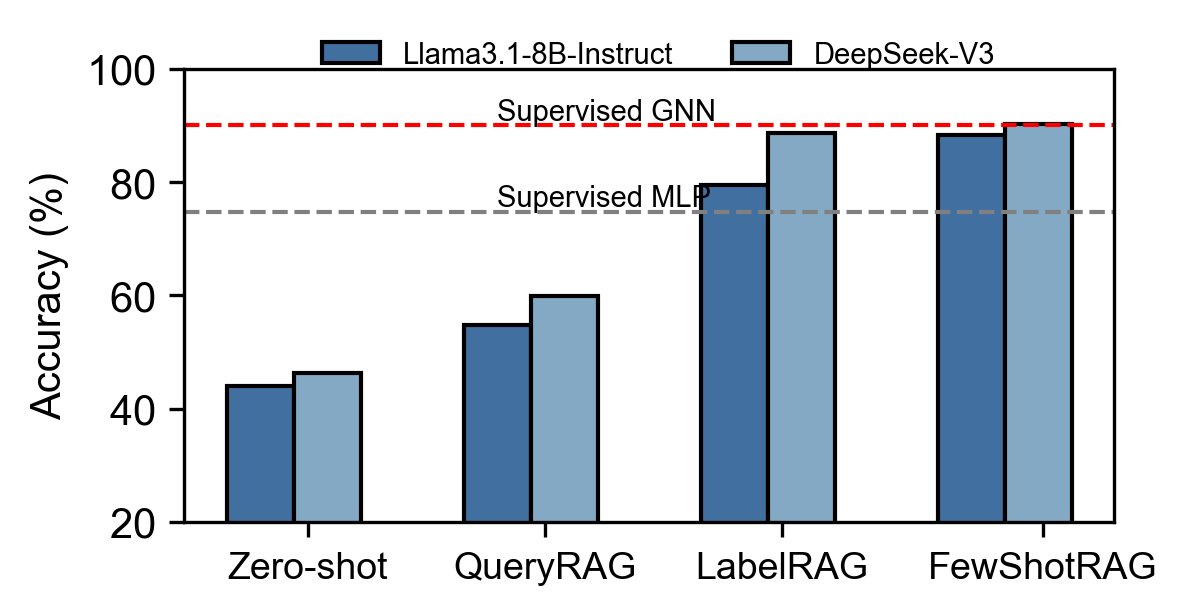}}
  \subfigure[History]{\includegraphics[width=0.16\linewidth]{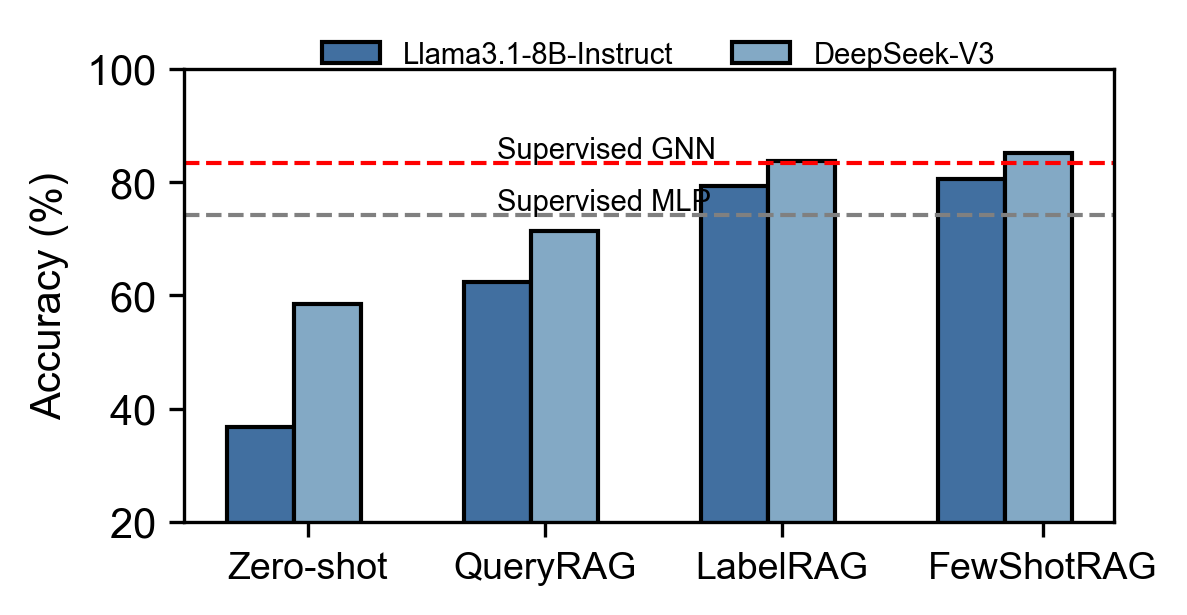}}
  \subfigure[Children]{\includegraphics[width=0.16\linewidth]{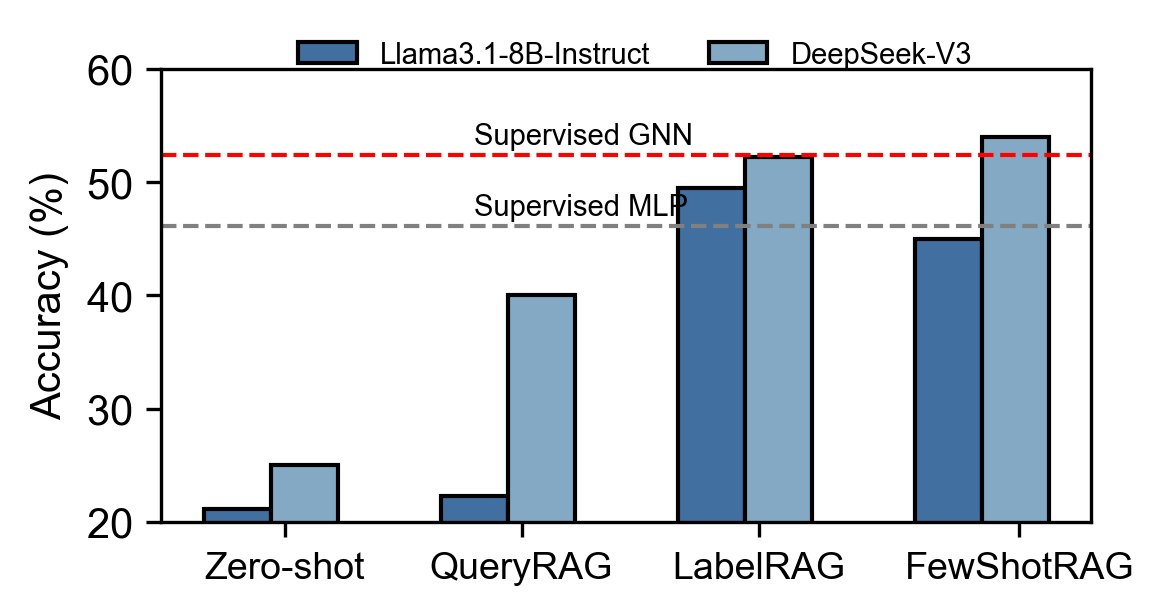}}
  \subfigure[Photo]{\includegraphics[width=0.16\linewidth]{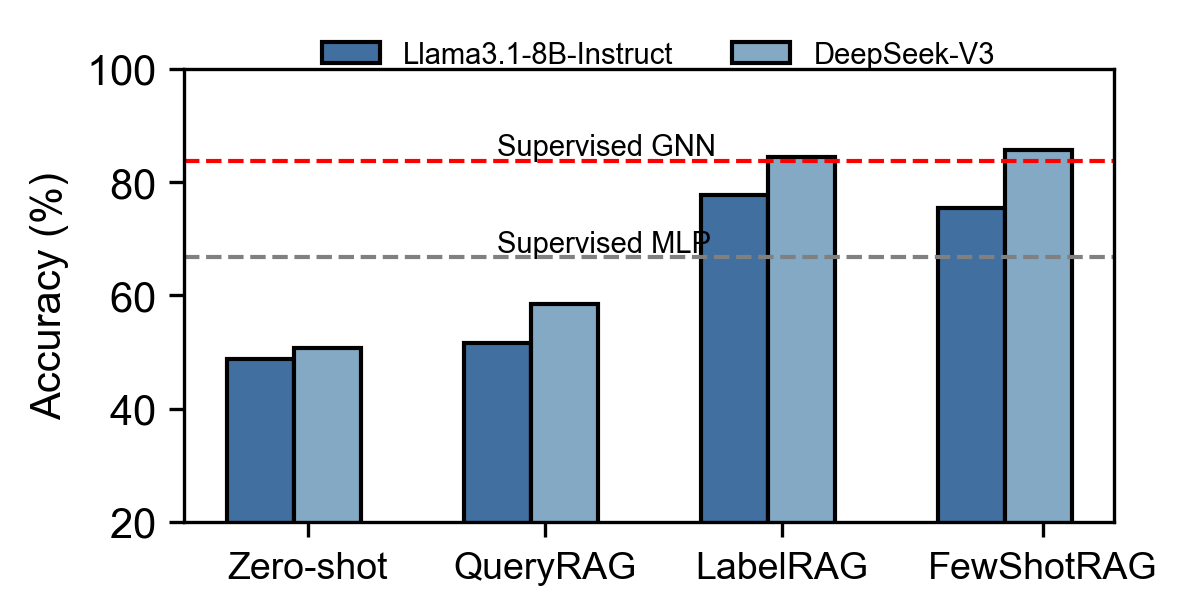}}
  \subfigure[Computers]{\includegraphics[width=0.16\linewidth]{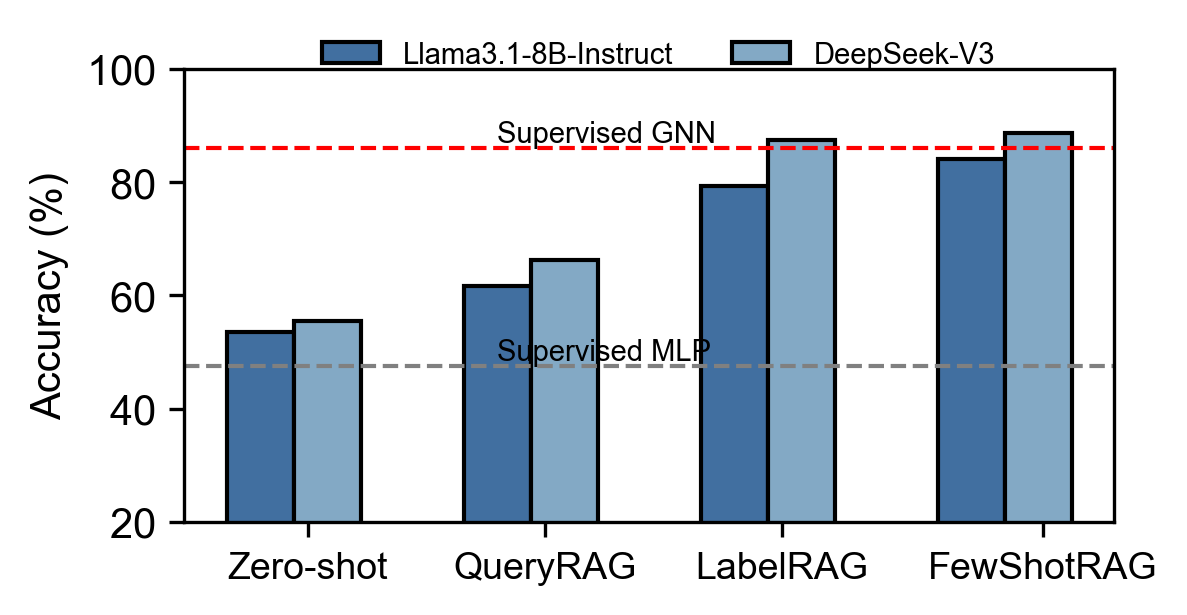}}
  \subfigure[arxiv]{\includegraphics[width=0.16\linewidth]{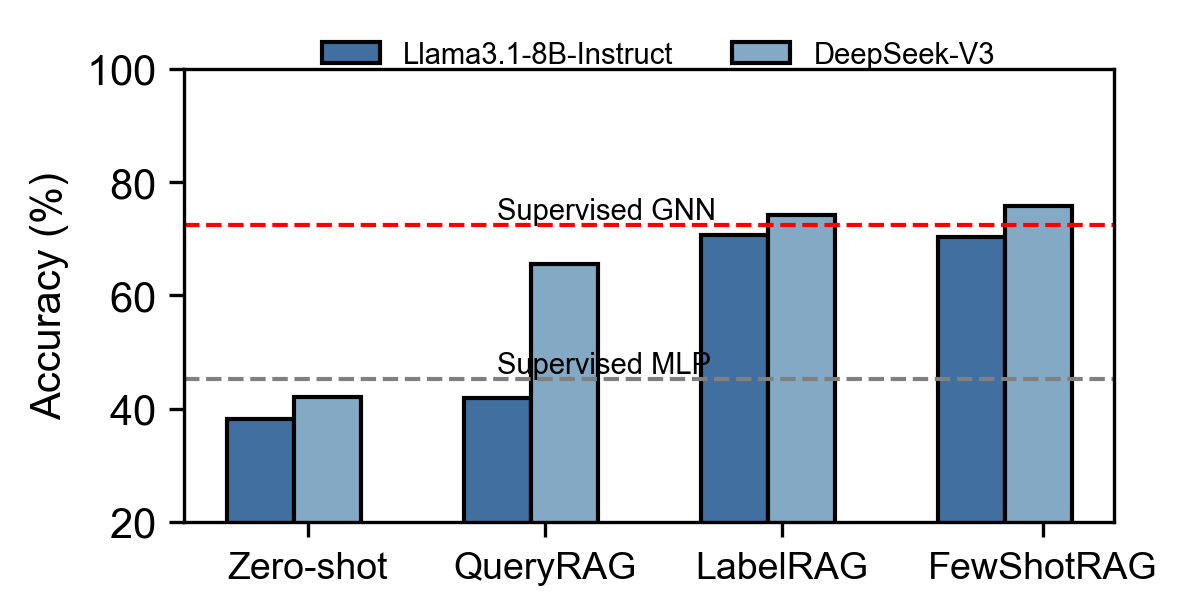}}
  \caption{In-context performance of \qrag, \lrag, and \frag compared to the zero-shot performance of \textsc{Llama-3.1-8B-Instruct} and \textsc{DeepSeek-V3}.}
  \label{fig:icl}
\end{figure*}

\begin{table*}[h!]
  \centering
  \footnotesize
  \begin{tabular}{lcccccccc}
    \toprule
                                   & Cora            & Pubmed          & Fitness                  & History         & Children        & Photo                    & Computers       & arxiv                   \\
    \midrule
    \midrule
    MLP                            & 55.6$_{\pm0.1}$ & 68.1$_{\pm0.5}$ & 78.8$_{\pm0.3}$          & 77.4$_{\pm0.3}$ & 47.1$_{\pm0.4}$ & 56.2$_{\pm0.2}$          & 64.7$_{\pm0.4}$ & 46.5$_{\pm0.2}$         \\
    GCN~\cite{gcn}                 & 81.3$_{\pm0.7}$ & 79.5$_{\pm0.3}$ & 90.0$_{\pm0.2}$          & 83.3$_{\pm0.8}$ & 52.4$_{\pm1.1}$ & 83.8$_{\pm0.2}$          & 86.1$_{\pm0.4}$ & 72.5$_{\pm0.2}$         \\
    GAT~\cite{gat}                 & 82.1$_{\pm0.9}$ & 79.1$_{\pm0.6}$ & \second{91.5}$_{\pm1.0}$ & 83.4$_{\pm1.2}$ & 53.1$_{\pm0.2}$ & 83.3$_{\pm0.2}$          & 86.6$_{\pm0.7}$ & 72.7$_{\pm0.3}$         \\
    GraphSAGE~\cite{graphsage}     & 78.6$_{\pm0.5}$ & 78.9$_{\pm0.6}$ & 89.8$_{\pm0.1}$          & 82.4$_{\pm0.8}$ & 52.0$_{\pm1.0}$ & 82.7$_{\pm0.7}$          & 84.4$_{\pm0.5}$ & 72.9$_{\pm0.5}$         \\
    RevGAT~\cite{revgat}           & 81.9$_{\pm0.2}$ & 80.1$_{\pm0.9}$ & \first{91.6}$_{\pm0.7}$  & 83.7$_{\pm0.7}$ & 52.9$_{\pm0.6}$ & \second{84.8$_{\pm0.5}$} & 87.0$_{\pm1.2}$ & \third{74.5$_{\pm0.4}$} \\
    GLEM~\cite{glem}               & 82.9            & 85.7            & 88.9                     & 84.0            & 52.1            & 79.8                     & 80.2            & 74.7                    \\
    TAPE~\cite{tape}               & 83.1            & {86.5}          & 89.0                     & \first{85.8}    & \second{53.4}   & 82.1                     & 84.5            & \second{75.2}           \\
    PRODIGY~\cite{PRODIGY}         & 80.5            & 82.4            & 85.7                     & 84.3            & 50.9            & 80.5                     & 86.1            & 74.3                    \\
    LLaGA~\cite{llaga}             & \third{83.4}    & \third{86.8}    & 90.0                     & \third{84.9}    & \third{53.2}    & 83.1                     & \second{88.0}   & \second{75.2}           \\
    \midrule
    \multicolumn{9}{c}{\textsc{Llama3.1-8B-Instruct}}                                                                                                                                                        \\
    \midrule
    Zero-shot~\cite{zero_shot_cot} & 57.8            & 66.3            & 43.9                     & 36.8            & 21.1            & 48.8                     & 53.6            & 38.1                    \\
    One-shot~\cite{few_shot}       & 34.7            & 47.9            & 22.4                     & 35.5            & 21.0            & 33.8                     & 39.7            & 14.7                    \\
    Few-shot~\cite{few_shot}       & 34.5            & 64.1            & 34.1                     & 39.8            & 24.0            & 35.0                     & 41.8            & 16.4                    \\
    RAG~\cite{rag}                 & 57.2            & 65.3            & 50.6                     & 39.0            & 20.1            & 48.7                     & 54.4            & 40.5                    \\
    \rowcolor{gray!10} \qrag       & 60.9            & 69.8            & 54.8                     & 62.3            & 22.3            & 51.5                     & 61.7            & 41.8                    \\
    \rowcolor{gray!10} \lrag       & 80.6            & 78.8            & 79.5                     & 79.3            & 49.5            & 77.8                     & 79.3            & 70.7                    \\
    \rowcolor{gray!10} \frag       & 81.9            & 77.9            & 88.3                     & 80.6            & 45.0            & 75.5                     & 84.1            & 70.3                    \\
    \midrule
    \multicolumn{9}{c}{\textsc{DeepSeek-V3}}                                                                                                                                                                 \\
    \midrule
    Zero-shot~\cite{zero_shot_cot} & 68.5            & 70.1            & 46.2                     & 58.5            & 25.0            & 50.7                     & 55.4            & 42.1                    \\
    One-shot~\cite{few_shot}       & 66.2            & 69.3            & 44.3                     & 57.8            & 24.3            & 45.0                     & 53.5            & 41.0                    \\
    Few-shot~\cite{few_shot}       & 69.5            & 66.9            & 49.5                     & 61.0            & 28.1            & 48.5                     & 60.3            & 46.8                    \\
    RAG~\cite{rag}                 & 69.4            & 71.9            & 57.8                     & 70.2            & 32.5            & 52.4                     & 60.7            & 57.2                    \\
    \rowcolor{gray!10} \qrag       & 75.2            & 78.3            & 59.9                     & 71.4            & 40.0            & 58.4                     & 66.3            & 65.5                    \\
    \rowcolor{gray!10} \lrag       & \second{84.0}   & \first{88.5}    & 88.7                     & {83.8}          & 52.2            & \third{84.5}             & \third{87.4}    & {74.2}                  \\
    \rowcolor{gray!10} \frag       & \first{86.5}    & \second{88.0}   & \third{90.2}             & \second{85.1}   & \first{54.0}    & \first{85.7}             & \first{88.6}    & \first{75.8}            \\
    \bottomrule
  \end{tabular}
  \caption{Node classification results (\%) on eight text-attributed graphs. The top-3 best performances in each dataset are highlighted in \first{red}, \second{blue}, and \third{gray}.}
  \label{tab:node_clas}
\end{table*}

\section{Experiments}
In this section, we present a rigorous evaluation of general-purpose LLMs augmented with our proposed RAG frameworks. The experiments are divided into two main parts: a comparative analysis of \qrag, \lrag, and \frag against state-of-the-art GNNs and graph LLMs on the node classification task, and an investigation of how different configurations impact the performance of general-purpose LLMs.
Due to space limitations, we defer the detailed experimental settings to Appendix~\ref{sec:exp_setting} and the ablation studies to Appendix~\ref{sec:add_res}.

\subsection{Experimental configurations}

\paragraph{Datasets.}
We conducted experiments on eight text-attributed graph datasets: Cora, Pubmed~\cite{sen2008collective}, and the CS-TAG~\cite{cstag} benchmarks, which include Sports-Fitness, Ele-History, Ele-Computers, Books-Children, Books-History, and Ogbn-Arxiv-TA. These datasets span diverse domains, such as citation networks and e-commerce, and vary in sparsity and size, ranging from small to large scales. The dataset statistics are provided in Table~\ref{tab:dataset}.
Since the primary focus of our work is node classification based on textual attributes, we also report the average number of input tokens for the textual attributes of nodes in these graphs. For the Cora and Pubmed datasets, we follow the public splits from \cite{gcn}, using 20 samples per class for training and 1,000 labeled examples for evaluation. For the CS-TAG datasets, we adopt a low-label-rate split, allocating 10\% of the data for training, 10\% for validation, and 80\% for testing.

\paragraph{Large language models.}
Our goal is to explore the in-context learning capabilities of LLMs. To this end, we evaluate a range of advanced language models, including \textsc{Llama-3.1-8B-Instruct}~\cite{llama3}, \textsc{Qwen2.5-7B-Instruct}~\cite{qwen2.5}, \textsc{Gemma-2-9B}~\cite{gemma_2024}, \textsc{Phi-3.5-Mini-Instruct}~\cite{phi-3}, and \textsc{Mistral-7B-Instruct-v0.3}~\cite{mistral}. Additionally, we conduct experiments with \textsc{DeepSeek-V3}~\cite{deepseekv3}, a closed-source LLM, to assess its performance on the node classification task under in-context learning settings. Due to budget constraints, we did not conduct comprehensive experiments with other advanced models, such as \textsc{GPT-4o}.

\paragraph{Baselines.}
In our experimental evaluation, we include a diverse range of state-of-the-art methods to ensure a comprehensive assessment: (i) supervised baseline without graph structure: MLP; (ii) supervised message-passing GNNs, including GCN~\cite{gcn}, GAT~\cite{gat}, GraphSAGE~\cite{graphsage}, and RevGAT~\cite{revgat}; (iii) specialized graph LLMs, including GLEM~\cite{glem}, TAPE~\cite{tape}, PRODIGY~\cite{PRODIGY}, and LLaGA~\cite{llaga}; and (iv) in-context baselines, including zero-shot, few-shot~\cite{few_shot}, and RAG~\cite{rag}. Detailed descriptions of the baselines used in our experiments are provided in Appendix~\ref{sec:exp_setting}.

\paragraph{Evaluation.}
For the supervised baselines, we train the models using the standard training splits provided for each dataset and evaluate their performance on the corresponding test splits. Accuracy is used as the performance metric, and we report the averaged scores and standard deviations across ten runs for MLP and GNNs. For LLM-based baselines, due to computational resource limitations, all results are averaged over three runs to account for variability in the few-shot sampling process.

\subsection{Results}

\paragraph{In-context performance.}
We investigate how our proposed RAG frameworks enhance the in-context learning capabilities of LLMs. Figure~\ref{fig:icl} presents the performance of \qrag, \lrag, and \frag with \textsc{LLAMA-3.1-8B-Instruct} and \textsc{DeepSeek-V3} across text-attributed graph benchmarks.
The results show that LLMs generally benefit from the additional contextual information provided by our RAG frameworks. In particular, LLMs with \lrag and \frag achieve significant performance improvements, outperforming supervised MLPs and, in some cases, matching or even surpassing supervised LLMs. Although GNNs still perform slightly better in certain cases, such as on the History dataset, the performance gap is small.
Overall, these findings indicate that our proposed RAG frameworks, especially \frag, serve as strong alternatives for graph-related tasks, enabling effective in-context learning without requiring any modifications to the LLMs.

\paragraph{Compared to baselines.}
We also present the results compared to various advanced baselines in Table~\ref{tab:node_clas}. Several key observations emerge from the results:
\begin{itemize}
  \item Graph-guided retrieval significantly improves LLM performance: All three RAG variants outperform standard zero-shot baselines (e.g., zero-shot, one-shot, and few-shot learning) on all datasets, demonstrating the effectiveness of graph-structured retrieval for augmenting LLM reasoning.
  \item Off-the-shelf-LLMs can match the performance of supervised or fine-tuned methods: With a sufficiently large LLM (e.g., \textsc{DeepSeek-V3}), the performance often matches or exceeds that of specialized graph LLMs like GLEM and TAPE. These results highlight the advantage of retrieval-augmented in-context learning, which effectively leverages graph structure as a source of context without requiring fine-tuning.
  \item \lrag is particularly effective in improving in-context learning: Simply incorporating label information from neighboring nodes (\lrag) results in a substantial accuracy gain compared to \qrag, suggesting that labeled neighborhood context is highly beneficial for LLM-based graph
        learning.
  \item \frag shows mixed results compared to \lrag: While \frag integrates both query and label information, it does not consistently outperform \lrag. This could be due to the increased input length and complexity, which may hinder LLMs from efficiently processing relevant information. This aligns with prior findings that LLMs struggle with long-context reasoning~\cite{lost_in_middle}.
\end{itemize}
Overall, these results highlight the potential of in-context learning for graph tasks while also revealing challenges related to input length sensitivity in LLMs.

\section{Related Work}

\paragraph{In-context learning.}
In-context learning (ICL) is a task adaptation strategy where a large language model (LLM) is provided with examples or demonstrations within the input prompt to guide its responses for various tasks~\cite{few_shot}. ICL enables LLMs to adapt to unseen tasks by observing a few demonstrations in their input, without requiring additional training data or fine-tuning~\cite{icl_survey}.
Since in-context demonstrations are crucial for ICL performance, various strategies have been proposed to improve demonstration selection, including retrieving semantically similar examples~\cite{LiuSZDCC22}, employing chain-of-thought reasoning~\cite{zero_shot_cot}, and decomposing tasks into subproblems using least-to-most prompting~\cite{ZhouSHWS0SCBLC23}. Additionally, research has shown that the effectiveness of ICL can be significantly improved through fine-tuning~\cite{WeiBZGYLDDL22} or adaptation during pretraining~\cite{metaicl}, even for smaller language models~\cite{SchickS21}. Recent work continues to refine demonstration selection strategies, further enhancing the adaptability and efficiency of in-context learning~\cite{LiuSZDCC22}.

\paragraph{Retrieval-augmented generation.}
Retrieval-augmented generation (RAG) is a context-based strategy that enhances LLMs by dynamically retrieving relevant information from a knowledge base or document corpus~\cite{rag}. Unlike ICL, which relies solely on in-prompt demonstrations, RAG provides external context to guide LLM responses, making it particularly effective for tasks that require up-to-date or domain-specific information where static prompts may be insufficient~\cite{rag_survey}.
RAG can also be integrated with ICL techniques, leveraging both retrieved knowledge and example-based reasoning to improve performance across diverse tasks~\cite{icl_survey}. More recently, several studies have explored the use of graph-aware retrieval mechanisms to improve RAG by providing structured contextual information~\cite{graphrag,gretriever}.

\paragraph{Graph neural networks.}
GNNs are powerful tools designed to operate on graph-structured data~\cite{message_passing}.
They have revolutionized the landscape by introducing the message-passing mechanism, where nodes iteratively aggregate information from their neighbors~\cite{chen2020survey}.
This paradigm has led to the development of several influential GNN architectures.
Popular GNN variants include graph convolutional networks (GCN)\cite{gcn}, graph attention networks (GAT)\cite{gat}, and scalable architectures such as GraphSAGE~\cite{graphsage} and RevGAT~\cite{revgat}, each advancing the message-passing paradigm in distinct ways.
While both GNNs and RAG leverage contextual information beyond the raw input, little work has explored the fundamental connections between them.

\section{Conclusion}
In this work, we establish a close connection between message-passing GNNs and RAG, which provides a holistic understanding of how GNNs process and leverage contextual information in graph data. By introducing \qrag, \lrag, and \frag, we extend the RAG paradigm to incorporate graph structure as inherent context, enabling effective in-context learning for LLMs without requiring fine-tuning or additional training. Extensive experimental evaluations demonstrate that our proposed RAG frameworks enable LLMs to achieve outstanding performance across multiple datasets, in some cases even surpassing state-of-the-art supervised GNNs and specialized graph LLMs. Beyond performance improvements, our work highlights the broader relationship between GNNs and RAG paradigms. By conceptualizing message passing as a recursive retrieval process, we bridge the gap between structured and unstructured learning, opening new avenues for applying LLMs to graph-based learning tasks.

\section*{Impact Statement}
This paper advances the field of machine learning by exploring the intersection of in-context learning and graph-based reasoning. Our proposed RAG frameworks enable LLMs to leverage structured graph data without requiring additional fine-tuning, expanding their applicability to knowledge-intensive domains.

While our work enhances LLMs’ reasoning capabilities, it also raises concerns about the reliability of retrieval-augmented methods, particularly when the retrieved context is noisy or biased. The effectiveness of these methods depends on the quality and relevance of the retrieved information, which may not always be trustworthy. Noisy, outdated, or adversarially manipulated sources can introduce errors, leading to unreliable model outputs. Future research should focus on developing adaptive retrieval mechanisms that evaluate and filter retrieved content based on reliability signals to improve robustness and trustworthiness.

\bibliography{example_paper}
\bibliographystyle{icml2025}

\newpage
\appendix
\onecolumn

\begin{table*}[ht]
  \centering
  \footnotesize
  \begin{tabular}{l|ccc}
    \toprule
    \textbf{Method}            & \textbf{Query} & \textbf{Context}                     & \textbf{Conditions}                                      \\ \midrule
    \midrule
    \textbf{Zero-shot}         & $q_i$          & -                                    & -                                                        \\
    \textbf{Few-shot}          & $q_i$          & $\{(q_1, r_1), \dots, (q_j, r_j)\}$  & -                                                        \\
    \textbf{RAG}               & $q_i$          & $\{c_1, \dots, c_j\}$                & $D(q_i, c_j) < \epsilon$                                 \\
    \midrule
    \textbf{GNN}               & $x_i$          & $\{x_1, \dots, x_j\}$                & $(v_i, v_j) \in \mathcal{E}$                             \\
    \textbf{Label Propagation} & $y_i$          & $\{y_1,\dots, y_j\}$                 & $(v_i, v_j) \in \mathcal{E}$                             \\
    \midrule
    \textbf{\qrag}             & $q_i$          & $\{q_1, \dots, q_j\}$                & $D(q_i, q_j) < \epsilon$ or $(v_i, v_j) \in \mathcal{E}$ \\
    \textbf{\lrag}             & $q_i$          & $\{r_1, \dots, r_j\}$                & $D(q_i, q_j) < \epsilon$ or $(v_i, v_j) \in \mathcal{E}$ \\
    \textbf{\frag}             & $q_i$          & $ \{(q_1, r_1), \dots, (q_j, r_j)\}$ & $D(q_i, q_j) < \epsilon$ or $(v_i, v_j) \in \mathcal{E}$
    \\ \bottomrule
  \end{tabular}
  \caption{Comparison of in-context prompting techniques and message passing approaches. $q$: input query, $r$: output response, $x$: node representation, $y$: ground-truth label, $c$: external corpus. For conventional RAG, we use the distance metric $D(\cdot)$ and a sufficiently small constant $\epsilon$ to represent the retrieval step.}
  \label{tab:comparison}
\end{table*}

\begin{table*}[!ht]
  \footnotesize
  \centering
  \begin{tabularx}{\textwidth}{lXc}
    \toprule
    \textbf{Prompt Name} & \textbf{Prompt Content}                                                                                                                                                                                                                                                                                                                                                                                                                                                                                                                             \\  \midrule
    \midrule
    Zero-shot            & \textbf{Task:} Classify the following text into one of the predefined categories: \promptfield{list of categories}. Make your decision based on the main topic and overall content of the text. If the text is ambiguous or does not clearly fit into any category, choose the closest match. Provide only the category name as the output. \newl \textbf{Text:} \promptfield{textual content} \newl \textbf{Answer:}
    \\ \midrule
    Few-shot             & \textbf{Task:} Classify the above text into one of the predefined categories: \promptfield{list of categories}. Use the provided few-shot examples to enhance your understanding of the topic and context for accurate classification. ...category name as the output. \newl \textbf{Examples:} \newl \textbf{Text}: \promptfield{few-shot text} \newl \textbf{Answer}: \promptfield{few-shot label} \newl …(more randomly sampled examples)… \newl \textbf{Text}: \promptfield{textual content} \newl \textbf{Answer:}
    \\ \midrule
    RAG                  & \textbf{Task:} Classify the above text into one of the predefined categories: \promptfield{list of categories}. Use the provided references to enhance your understanding of the topic and context for accurate classification. ...category name as the output. \newl \textbf{Reference:} \newl \textbf{Text}: \promptfield{retrieved text} \newl …(more retrieved texts)… \newl \textbf{Text}: \promptfield{textual content} \newl \textbf{Answer:}
    \\ \midrule
    \qrag                & \textbf{Task:} Classify the above text into one of the predefined categories: \promptfield{list of categories}. Use the provided references to enhance your understanding of the topic and context for accurate classification. ...category name as the output. \newl \textbf{Reference:} \newl \textbf{Text}: \promptfield{text of retrieved nodes} \newl …(more graph retrieved texts)… \newl \textbf{Text}: \promptfield{textual content} \newl \textbf{Answer:}                                                                                 \\ \midrule
    \lrag                & \textbf{Task:} Classify the above text into one of the predefined categories: \promptfield{list of categories}. Use the provided category of reference texts to enhance your understanding of the topic and context for accurate classification. ...category name as the output. \newl \textbf{Reference:} \newl \textbf{Answer}: \promptfield{category of retrieved nodes} \newl …(more graph retrieved labels)… \newl \textbf{Text}: \promptfield{textual content} \newl \textbf{Answer:}                                                         \\ \midrule
    \frag                & \textbf{Task:} Classify the above text into one of the predefined categories: \promptfield{list of categories}. Use the provided references and corresponding categories to enhance your understanding of the topic and context for accurate classification. ...category name as the output. \newl \textbf{Reference:} \newl \textbf{Text}: \promptfield{few-shot text} \newl \textbf{Answer}: \promptfield{few-shot label} \newl …(more graph-guided few-shot examples)… \newl \textbf{Text}: \promptfield{textual content} \newl \textbf{Answer:} \\ \bottomrule
  \end{tabularx}
  \caption{An illustration of prompts we use for in-context node classification on text-attributed graphs.}
  \label{tab:prompts}
\end{table*}

\begin{table*}[ht]
  \centering
  \small
  {
    \begin{tabular}{lcccccccc}
      \toprule
      \textbf{Dataset} & \textbf{\#Nodes} & \textbf{\#Edges} & \textbf{\#Classes} & \textbf{Avg. token length} & \textbf{Homophily} & \textbf{Avg. degree} \\
      \midrule
      \midrule
      Cora             & 2,708            & 10,556           & 7                  & 166                        & 0.81               & 3.90                 \\
      Pubmed           & 19,717           & 88,648           & 3                  & 370                        & 0.80               & 4.50                 \\
      Books-Children   & 76,875           & 1,554,578        & 24                 & 273                        & 0.42               & 30.24                \\
      Books-History    & 41,551           & 358,574          & 12                 & 297                        & 0.66               & 12.11                \\
      Ele-Computers    & 87,229           & 721,081          & 10                 & 113                        & 0.83               & 14.41                \\
      Ele-Photo        & 48,362           & 500,928          & 12                 & 183                        & 0.75               & 18.07                \\
      Sports-Fitness   & 173,055          & 1,773,500        & 13                 & 30                         & 0.90               & 17.45                \\
      Ogbn-Arxiv-TA    & 169,343          & 1,166,243        & 40                 & 222                        & 0.66               & 13.67                \\
      \bottomrule
    \end{tabular}
  }
  \caption{Datasets statistics.}
  \label{tab:dataset}
\end{table*}

\section{Discussion}
\paragraph{Comparison.}
Table~\ref{tab:comparison} compares different in-context learning techniques and message-passing approaches, illustrating their query inputs, contextual information, and retrieval conditions. Traditional in-context learning methods, such as zero-shot and few-shot learning, rely on either no context or a small set of demonstrations, while RAG retrieves external corpus elements based on a similarity metric. In contrast, GNNs and label propagation utilize graph structure to aggregate node features or labels from neighboring nodes. Our proposed frameworks, \qrag, \lrag, and \frag, bridge the gap between these paradigms by incorporating graph-guided retrieval as context for LLMs. Specifically, \qrag retrieves query information from neighboring nodes, \lrag retrieves labels, and \frag combines both, enhancing the in-context learning capabilities of LLMs. By leveraging both graph connectivity and semantic similarity, these frameworks provide a structured yet flexible approach to improving LLM performance on graph-based tasks.

\paragraph{Limitations.}
Despite the promising results, our work has certain limitations. First, our approach assumes homophily in the graph structure, which may not generalize well to heterophilic graphs, where nodes with dissimilar attributes or labels are more likely to be connected. Second, our experiments primarily focus on static graphs, leaving the performance on dynamic or temporal graphs unexplored. Finally, while GNNs enhance the understanding of a query node by recursively aggregating information from its local neighborhood, RAG typically retrieves only one-hop context for LLMs, which may limit their ability to handle complex multi-hop reasoning tasks~\cite{hotpotqa}.

\paragraph{Future work.}
For future work, we aim to investigate the in-context learning performance of LLMs in more challenging settings, where graphs may be dynamic, noisy, incomplete, or entirely unavailable. Additionally, we plan to extend our methods to larger and sparser graphs by exploring retrieval mechanisms that capture long-range node relationships beyond homophily.
Beyond graph learning, we also plan to explore the broader applicability of our frameworks to natural language processing tasks. Specifically, we seek to apply our methods to structured knowledge retrieval, reasoning over knowledge graphs, and few-shot adaptation to domain-specific problems. These extensions would further demonstrate the versatility of RAG-based approaches and their ability to integrate structured and unstructured learning paradigms effectively.

\section{Experiment settings}
\label{sec:exp_setting}

\subsection{Datasets}
We provide detailed descriptions of the datasets used in the main experiments as follows:
\begin{itemize}
  \item \textbf{Cora and Pubmed}~\cite{sen2008collective} are widely used citation graph datasets in graph learning research. We use the commonly adopted versions\cite{gcn} and collected the missing textual attributes for these datasets to ensure consistency with other datasets.
  \item \textbf{Ogbn-Arxiv-TA (arxiv)}~\cite{cstag} is derived from the citation dataset Ogbn-Arxiv\cite{hu2020ogb}. The task involves predicting the categories of the papers, formulated as a 40-class classification problem. The text attributes for each paper node are extracted from its title and abstract in Ogbn-Arxiv.
  \item \textbf{Books-Children and Books-History}~\cite{cstag} are extracted from the Amazon-Books dataset. Books-Children contains items with the second-level label Children', while Books-History contains items with the second-level label History’. The nodes represent books, and edges indicate that two books are frequently co-purchased or co-viewed. The labels correspond to the third-level categories of the books. The title and description of each book are used as the text attributes for the nodes.
  \item \textbf{Ele-Computers and Ele-Photo}~\cite{cstag} are extracted from the Amazon-Electronics dataset. Ele-Computers contains items with the second-level label Computers', and Ele-Photo contains items with the second-level label Photo’. Nodes represent electronics-related products, and an edge indicates frequent co-purchase or co-viewing between two products.
  \item \textbf{Sports-Fitness}~\cite{cstag} is extracted from the Amazon-Sports dataset. It consists of items with the second-level label `Fitness’. Nodes represent fitness-related items, and an edge between two items indicates frequent co-purchase or co-viewing.
\end{itemize}
For the Cora and Pubmed datasets, we use the original splits from \cite{gcn}, which sample 20 nodes per class for the training set and 500/1000 nodes for validation/test, respectively. For the arxiv dataset, we adopt the public split provided in \cite{cstag,hu2020ogb}. For the remaining datasets, we use a random 10\%/10\%/80\% split for train/validation/test to align with the strict settings typically encountered in real-world scenarios.

\subsection{Baselines}
We provide detailed descriptions of the baselines used in the main experiments as follows:

(i) Supervised GNNs:
\begin{itemize}
  \item \textbf{GCN}~\cite{gcn}: GCN is a foundational GNN model that applies convolution operations on graph-structured data, aggregating information from neighboring nodes.
  \item \textbf{GAT}~\cite{gat}: GAT introduces attention mechanisms to learn the importance of neighboring nodes, enabling more expressive and context-sensitive node feature aggregation.
  \item \textbf{GraphSAGE}~\cite{graphsage}: GraphSAGE uses a sampling-based neighborhood aggregation approach to efficiently learn node embeddings, particularly for large-scale graphs.
  \item \textbf{RevGAT}~\cite{revgat}: RevGAT combines reversible connectivity with a deep network architecture, enabling memory-efficient and scalable training of GNNs with deeper layers.
\end{itemize}

(ii) Graph LLMs:
\begin{itemize}
  \item \textbf{GLEM}~\cite{glem}: GLEM is a variational Expectation Maximization framework that trains LLMs and GNNs jointly on a graph. We adopt DeBERTa~\cite{deberta} as the language model for fine-tuning, following the original implementation.
  \item \textbf{TAPE}~\cite{tape}: TAPE is an LLM-enhanced framework that leverages textual explanations generated by LLMs to enhance the performance of downstream GNNs. In our experiments, we use the original explanations provided by the authors for the Cora, Pubmed, and arXiv datasets, and we generate explanations for the remaining datasets using \textsc{Llama3.1-8B-Instruct}. The downstream GNN used is a two-layer GCN, and the encoding model for the textual explanations is DeBERTa~\cite{deberta}.
  \item \textbf{PRODIGY}~\cite{PRODIGY}: PRODIGY is a pretraining framework that formulates in-context learning over graphs with a novel prompt graph representation. Following the original implementation, we pretrain PRODIGY on arXiv and evaluate its in-context performance on other datasets.
  \item \textbf{LLaGA}~\cite{llaga}: LLaGA is an LLM-based graph learning framework that introduces a versatile linear projector to seamlessly bridge graph structures with the token space understood by LLMs. We implement LLaGA using Vicuna-7B-v1.5-16K~\cite{vicuna} as the foundational base model and SimTeg~\cite{simteg} as the default text-encoding model.
\end{itemize}

(iii) In-context baselines:
\begin{itemize}
  \item \textbf{Few-shot}~\cite{few_shot}: Few-shot learning provides a small number of labeled examples as context for the language model to perform downstream tasks without fine-tuning. In our experiments, we randomly draw $K = \text{deg}(v_i)$ samples from the graph as in-context examples for node $v_i$.
  \item \textbf{One-shot}~\cite{few_shot}: One-shot learning provides a single labeled example to guide the language model in solving tasks with minimal supervision, equivalent to $K = 1$ in few-shot learning.
  \item \textbf{Zero-shot}~\cite{zero_shot_cot}: Zero-shot learning evaluates the language model using only task instructions without any labeled examples or demonstrations. This setting reflects the basic capability of LLMs to solve graph learning tasks.
  \item \textbf{RAG}~\cite{rag}: RAG combines pre-trained language models with a retriever mechanism to fetch relevant context for in-context learning and improve task performance. We use BERT~\cite{bert} as the embedding model and the FAISS library~\cite{faiss} for efficient similarity search.
\end{itemize}
(iv) We provide the download links for the LLM on Hugging Face and ModelScope below:
\begin{itemize}
  \item \textsc{Llama3.1-8b-Instruct}:
        \begin{itemize}
          \item Hugging Face: \url{https://huggingface.co/meta-llama/Llama-3.1-8B-Instruct}
          \item ModelScope: \url{https://www.modelscope.cn/LLM-Research/Meta-Llama-3.1-8B-Instruct.git}
        \end{itemize}
  \item \textsc{Qwen2.5-7b-Instruct}:
        \begin{itemize}
          \item Hugging Face: \url{https://huggingface.co/Qwen/Qwen2.5-7B-Instruct}
          \item ModelScope: \url{https://www.modelscope.cn/qwen/Qwen2.5-7B-Instruct.git}
        \end{itemize}
  \item \textsc{Gemma-2-9B-it}:
        \begin{itemize}
          \item Hugging Face: \url{https://huggingface.co/google/gemma-2-9b-it}
          \item ModelScope: \url{https://www.modelscope.cn/LLM-Research/gemma-2-9b-it.git}
        \end{itemize}
  \item \textsc{Phi-3.5-Mini-Instruct}:
        \begin{itemize}
          \item Hugging Face: \url{https://huggingface.co/microsoft/Phi-3.5-mini-instruct}
          \item ModelScope: \url{https://www.modelscope.cn/LLM-Research/Phi-3.5-mini-instruct.git}
        \end{itemize}
  \item \textsc{Mistral-7B-Instruct-v0.3}:
        \begin{itemize}
          \item Hugging Face: \url{https://huggingface.co/mistralai/Mistral-7B-Instruct-v0.3}
          \item ModelScope: \url{https://www.modelscope.cn/LLM-Research/Mistral-7B-Instruct-v0.3.git}
        \end{itemize}
\end{itemize}

\subsection{Implementations}
For few-shot learning, we evaluate each example in the test set by randomly selecting  $K$  examples from the task’s training set as conditioning examples. In the zero-shot setting, we use a natural language prompt, and for  $K = 0$, no demonstrations are provided. The same evaluation protocol is applied across all LLM backbones to ensure a fair comparison.
For \qrag, \lrag, and \frag, we evaluate them in few-shot in-context learning setups without additional fine-tuning of the LLMs. In these setups, textual context and/or label information obtained from the \textit{one-hop} neighbors of each node are incorporated into the prompts. Our implementation is based on the ms-swift~\cite{swift} and transformers~\cite{transformers} libraries. The experiments were conducted on a Linux system with 8 NVIDIA 4090 GPUs, each with 24GB of memory. We present the prompts used for in-context node classification on text-attributed graphs in Table~\ref{tab:prompts}.

\section{Additional experiment results}
\label{sec:add_res}

\begin{figure}
  \centering
  \subfigure[Cora]{\includegraphics[width=0.23\linewidth]{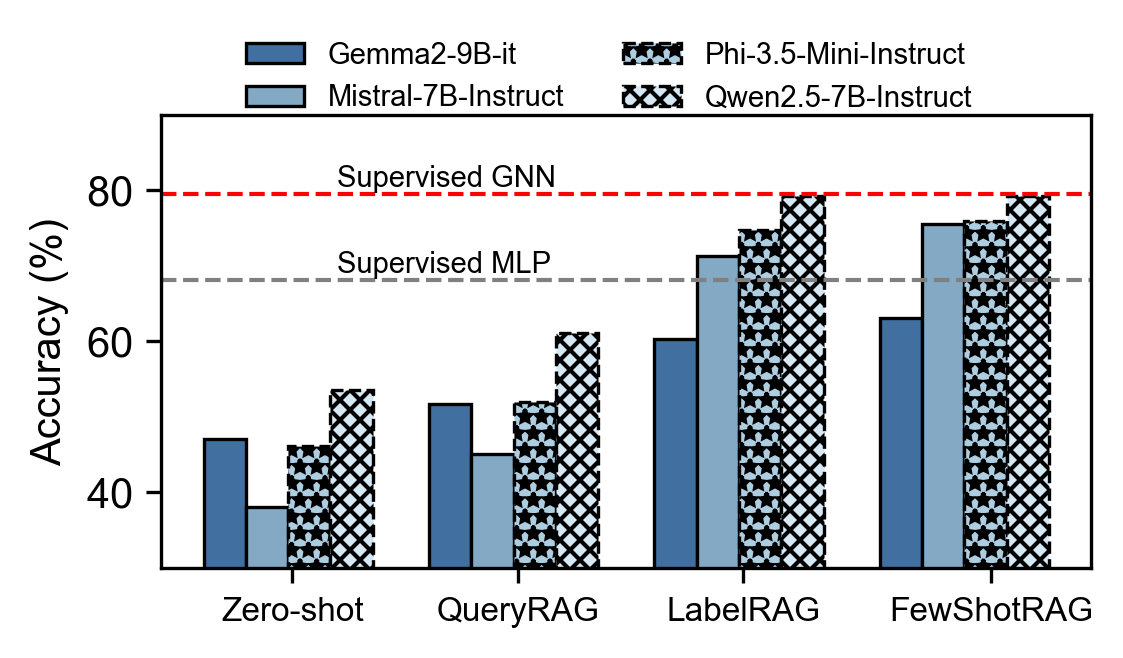}}
  \subfigure[Pubmed]{\includegraphics[width=0.23\linewidth]{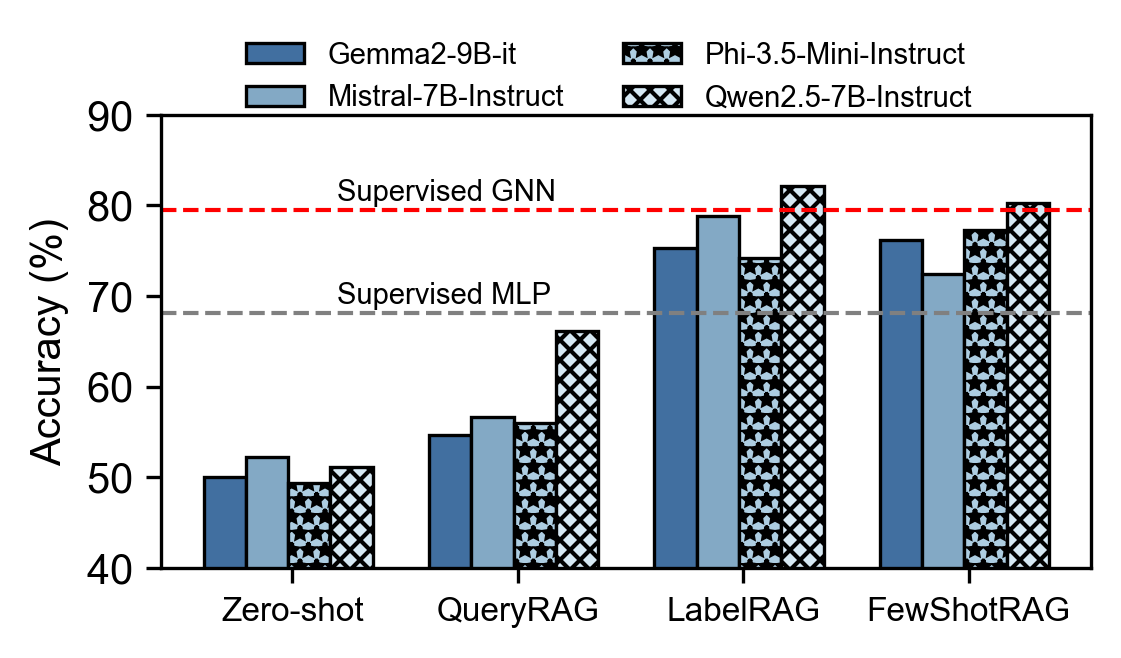}}
  \subfigure[Fitness]{\includegraphics[width=0.23\linewidth]{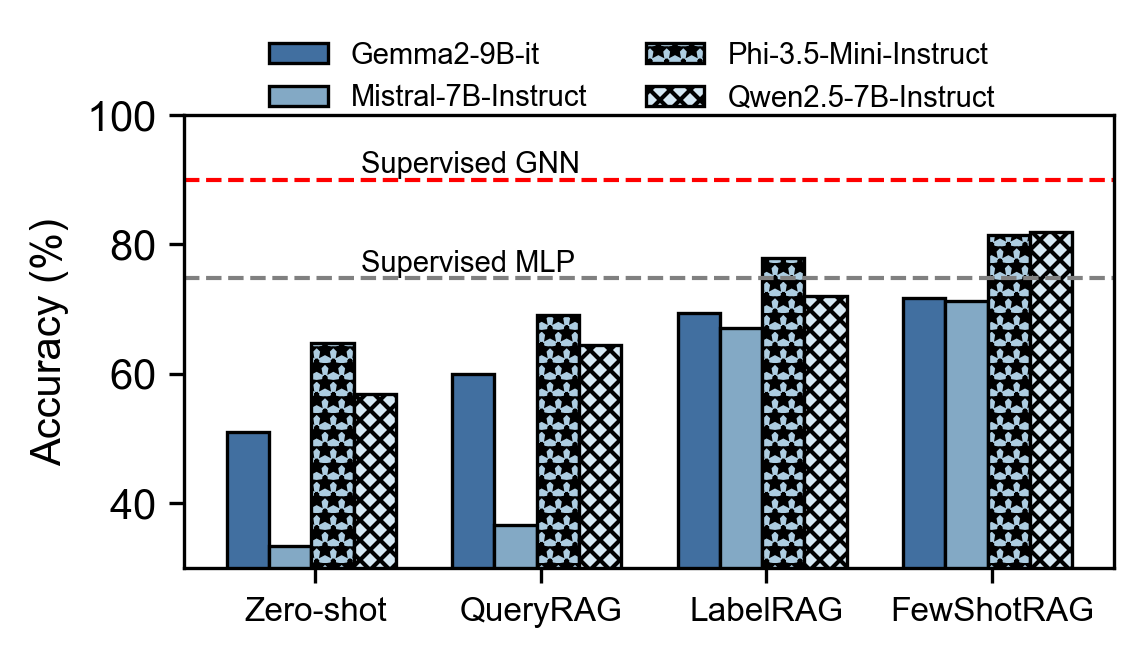}}
  \subfigure[History]{\includegraphics[width=0.23\linewidth]{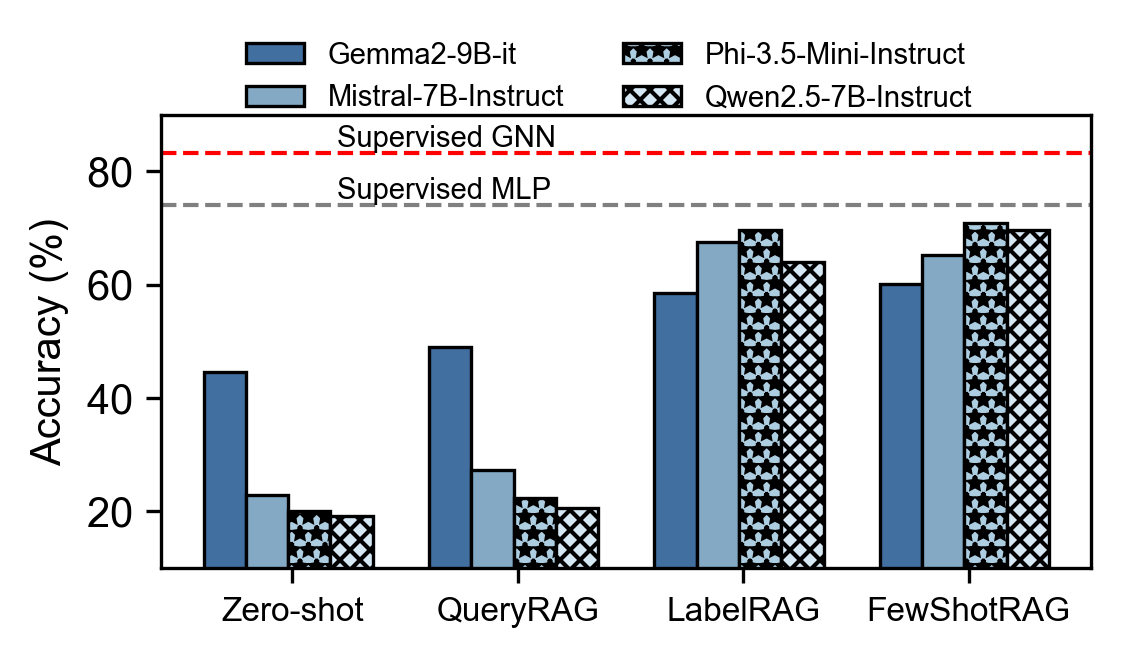}}
  \subfigure[Children]{\includegraphics[width=0.23\linewidth]{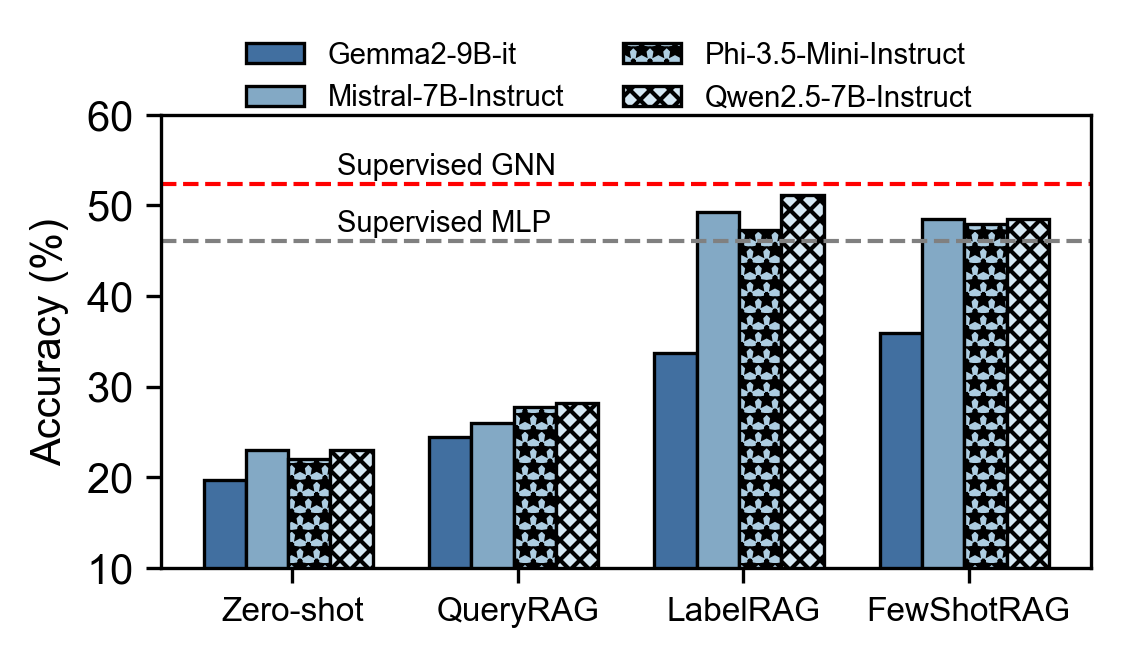}}
  \subfigure[Photo]{\includegraphics[width=0.23\linewidth]{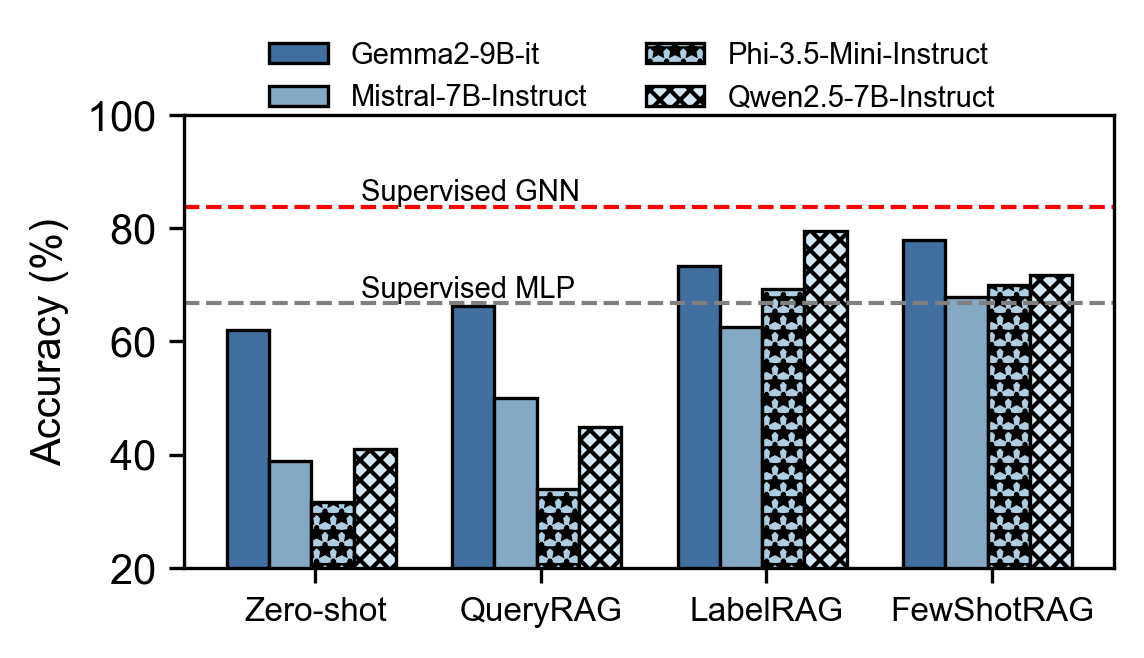}}
  \subfigure[Computers]{\includegraphics[width=0.23\linewidth]{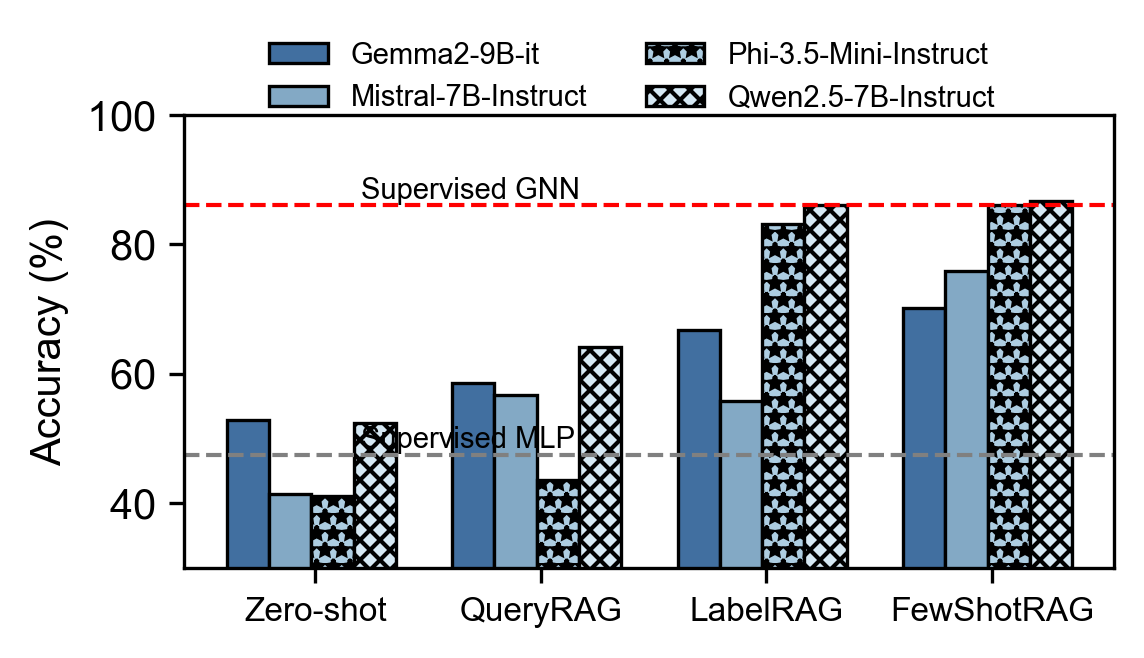}}
  \subfigure[Arxiv]{\includegraphics[width=0.23\linewidth]{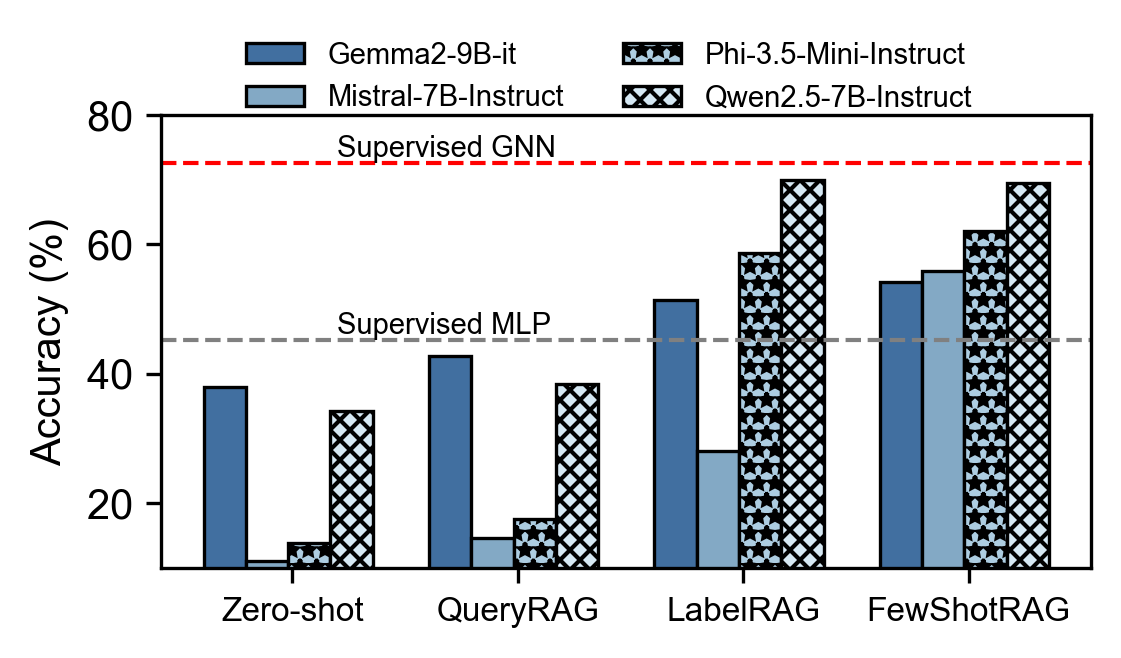}}
  \caption{Ablation results on the use of different LLMs.}
  \label{fig:llm}
\end{figure}

\paragraph{Different LLM backbones.}
To validate the generality of our proposed RAG frameworks, we evaluate them using additional LLMs, including \textsc{Qwen2.5-7B-Instruct}\cite{qwen2.5}, \textsc{Gemma-2-9B}\cite{gemma_2024}, \textsc{Phi-3.5-Mini-Instruct}\cite{phi-3}, and \textsc{Mistral-7B-Instruct-v0.3}\cite{mistral}. The experimental results are presented in Figure~\ref{fig:llm}.
Across eight datasets, all proposed RAG frameworks (\qrag, \lrag, and \frag) consistently improve performance across different LLM backbones. Among them, \frag achieves the highest accuracy across all models, highlighting the advantage of incorporating both query and label information. These results confirm that our RAG frameworks are versatile and can effectively adapt to various LLM backbones, providing a scalable solution for graph-related tasks while leveraging the strengths of diverse pre-trained models.

\begin{figure}
  \centering
  \subfigure[Cora]{\includegraphics[width=0.23\linewidth]{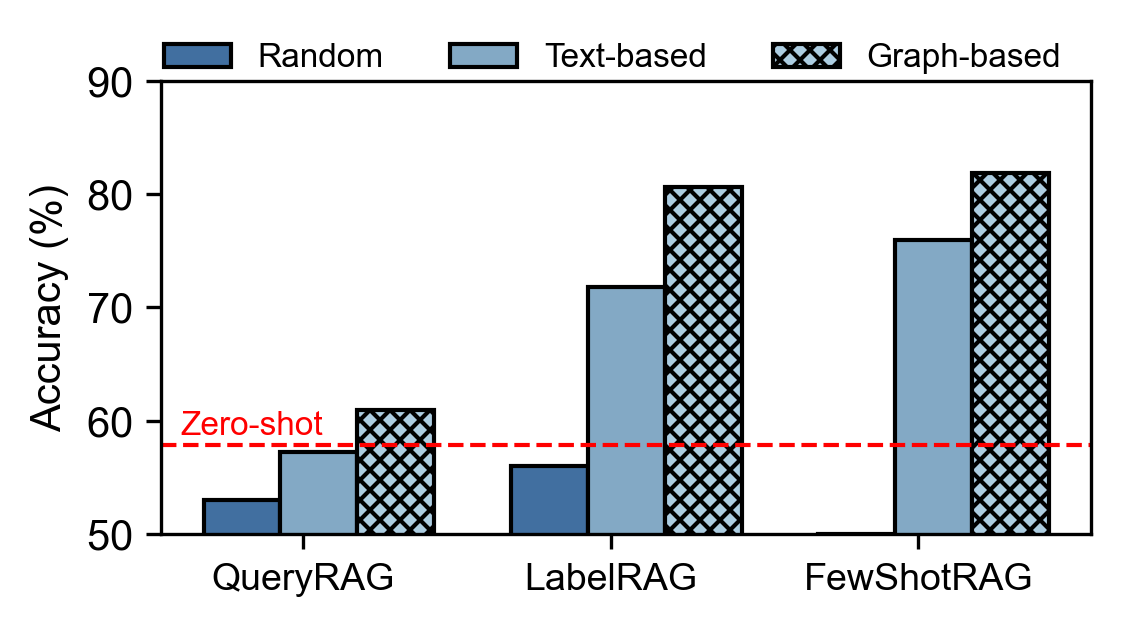}}
  \subfigure[Pubmed]{\includegraphics[width=0.23\linewidth]{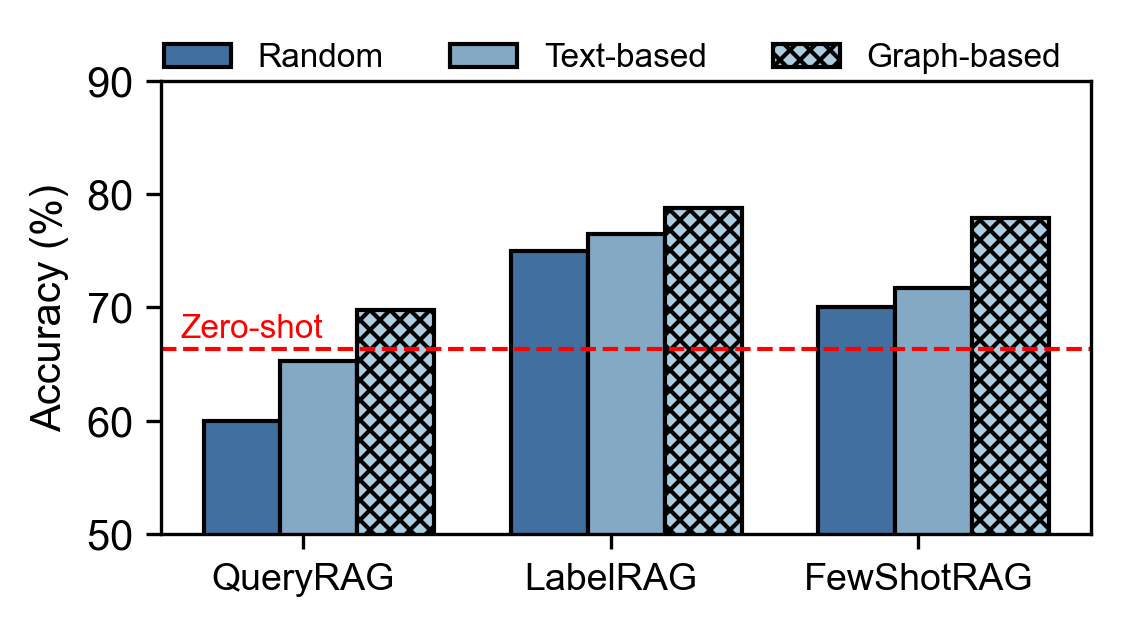}}
  \subfigure[Fitness]{\includegraphics[width=0.23\linewidth]{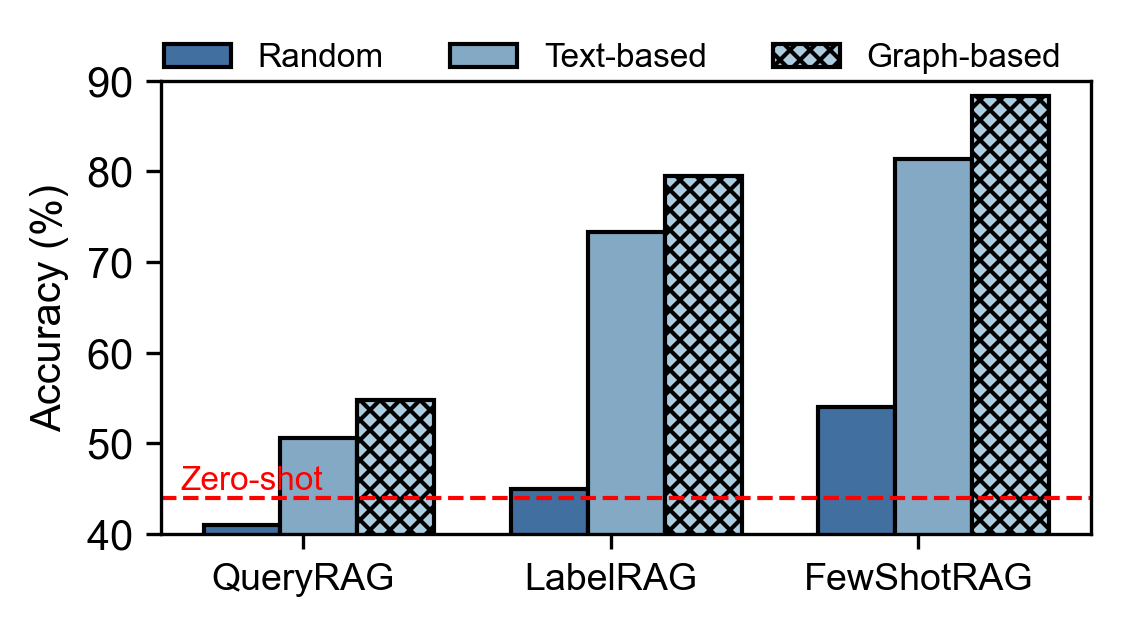}}
  \subfigure[History]{\includegraphics[width=0.23\linewidth]{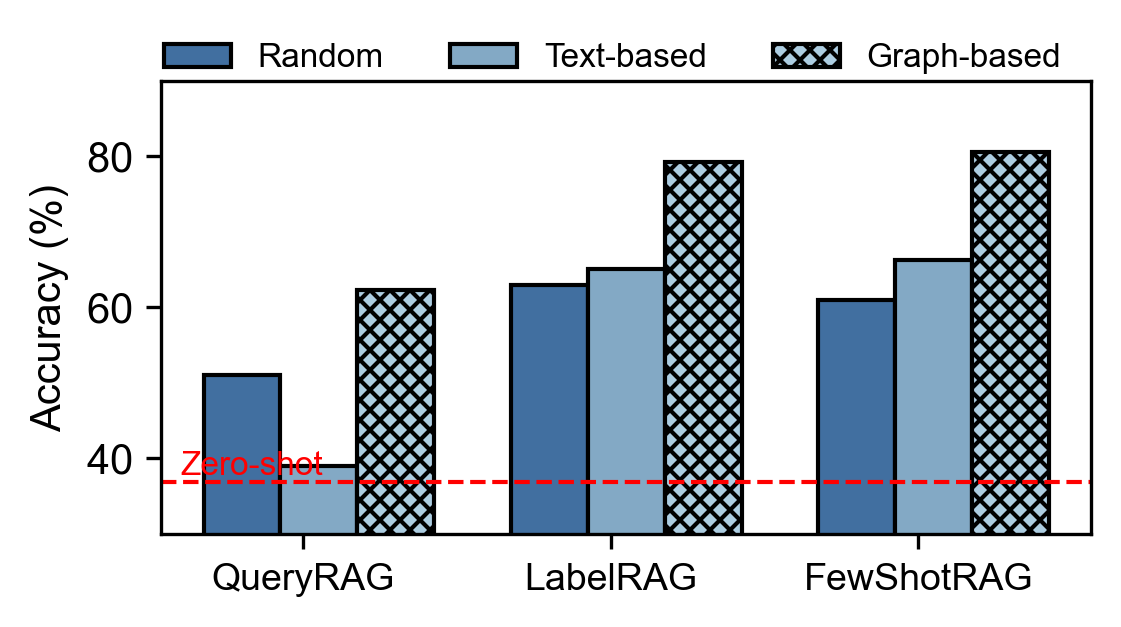}}
  \subfigure[Children]{\includegraphics[width=0.23\linewidth]{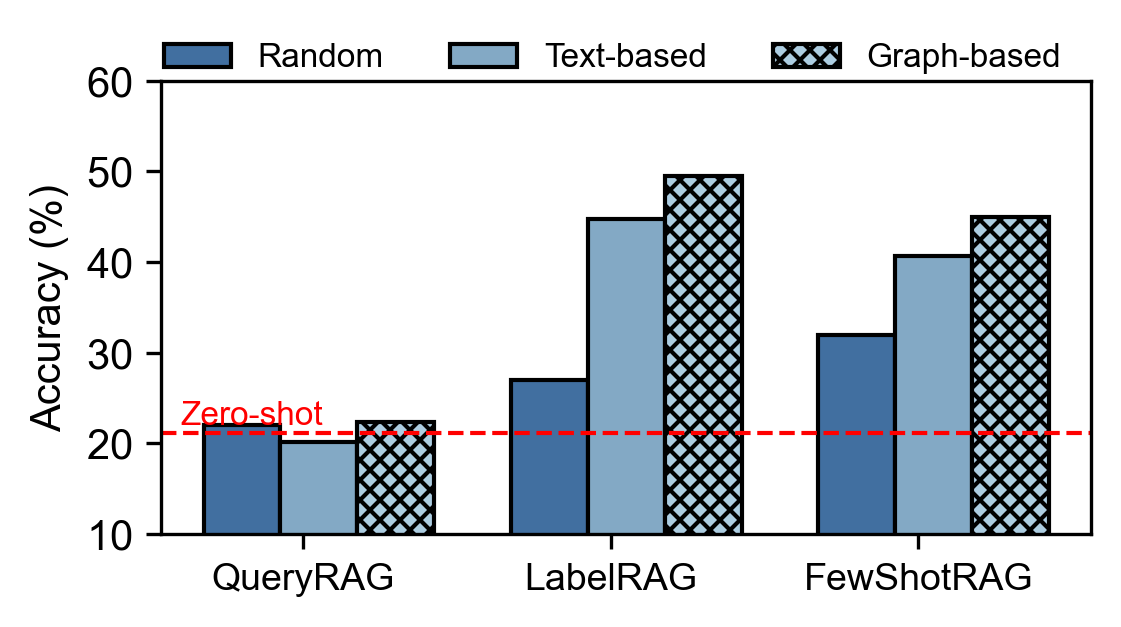}}
  \subfigure[Photo]{\includegraphics[width=0.23\linewidth]{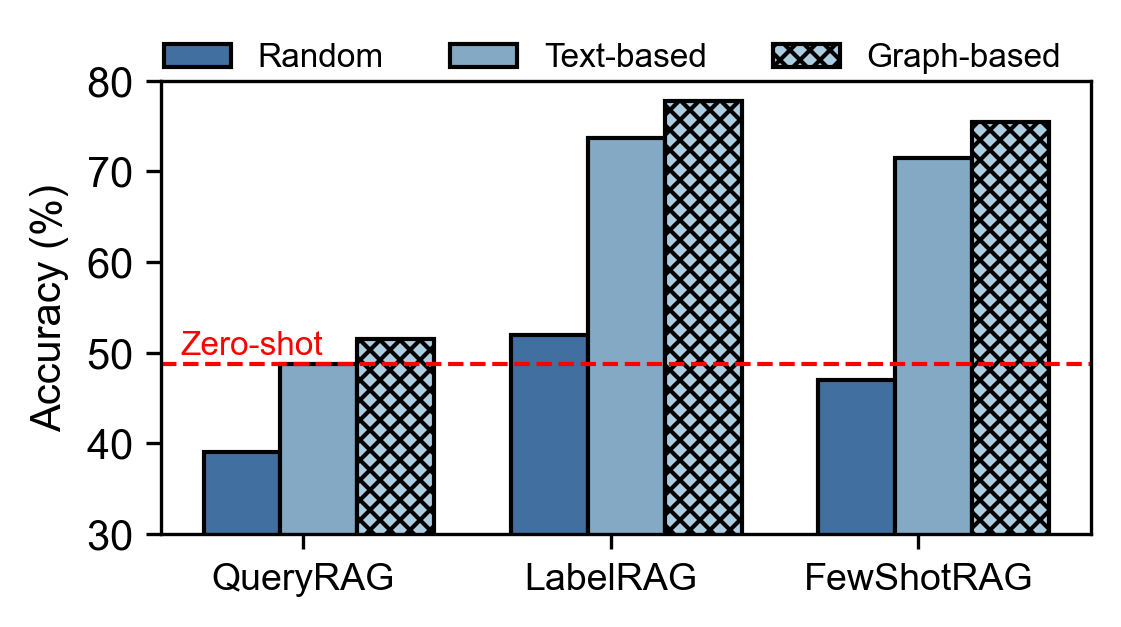}}
  \subfigure[Computers]{\includegraphics[width=0.23\linewidth]{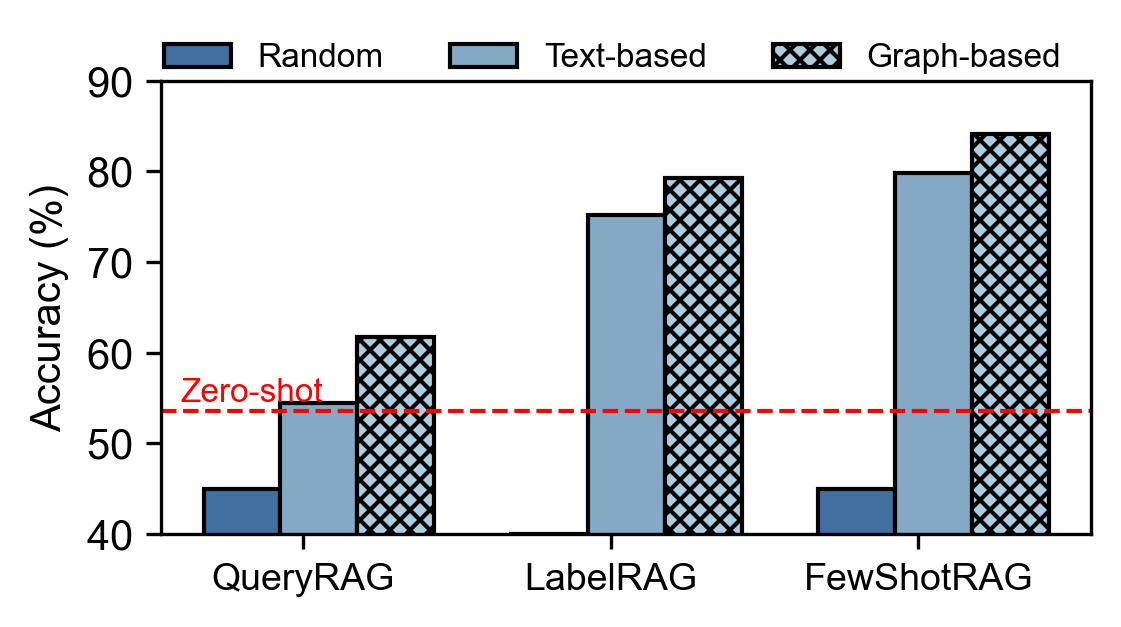}}
  \subfigure[arxiv]{\includegraphics[width=0.23\linewidth]{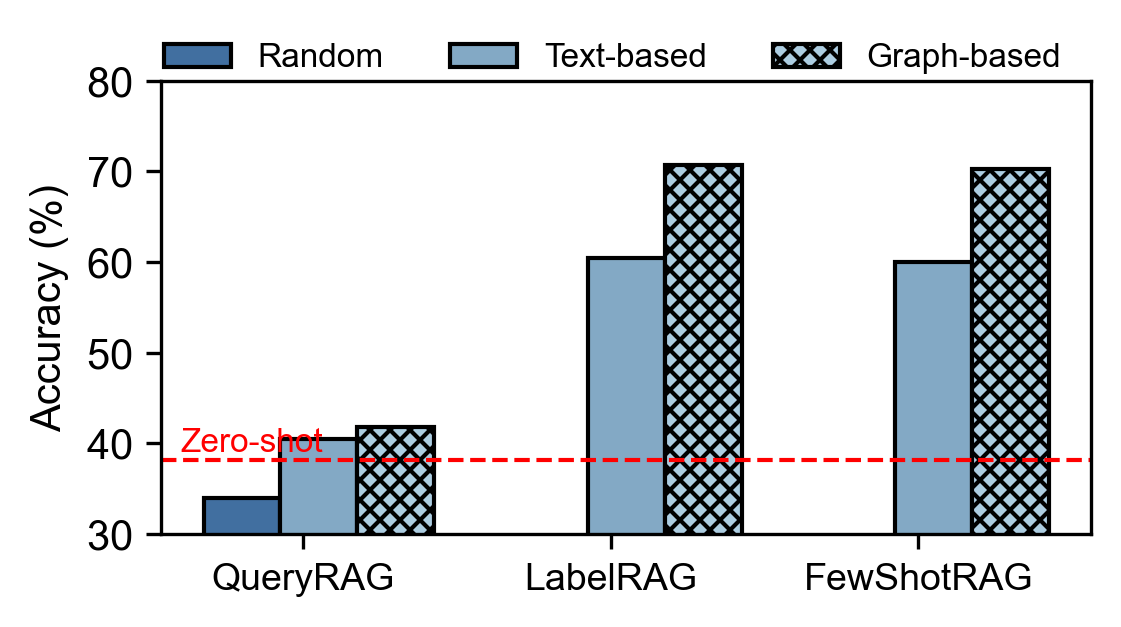}}
  \caption{Ablation results on different retrieval mechanisms.}
  \label{fig:retriever}
\end{figure}

\paragraph{Retriever.}
\qrag, \lrag, and \frag were initially proposed as RAG frameworks that utilize graph structure as inherent context, eliminating the need for additional retrieval efforts. Since \textit{context} is crucial for the in-context learning capabilities of LLMs, we perform ablation experiments on the retriever in our proposed RAG frameworks, comparing random-based, text-based, and graph-based retrieval methods.
For text-based retrieval, we follow the standard RAG protocol, retrieving textual neighbors based on node attributes rather than relying on graph structure. In this setup, the retriever leverages dense text embeddings to locate relevant documents~\cite{DPR}. These embeddings are generated using a pre-trained BERT model~\cite{bert}, which captures semantic similarity between nodes’ textual attributes. The retrieved context is then incorporated into input prompts for LLMs. To ensure consistency, the number of retrieved neighbors matches the number of graph-based neighbors for each node.
The ablation results, shown in Figure~\ref{fig:retriever}, evaluate the impact of replacing graph structure with random-based and text-based retrieval on the all datasets. We use \textsc{Llama3.1-8B-Instruct} as the LLM model. As observed, graph-based retrieval consistently outperforms both text-based and random-based retrieval across \qrag, \lrag, and \frag, aligning with findings from GraphRAG~\cite{graphrag}. These results highlight the importance of graph structure as a source of inherent context and its critical role in enhancing the in-context learning capabilities of LLMs.

\begin{figure}
  \centering
  \subfigure[Cora]{\includegraphics[width=0.25\linewidth]{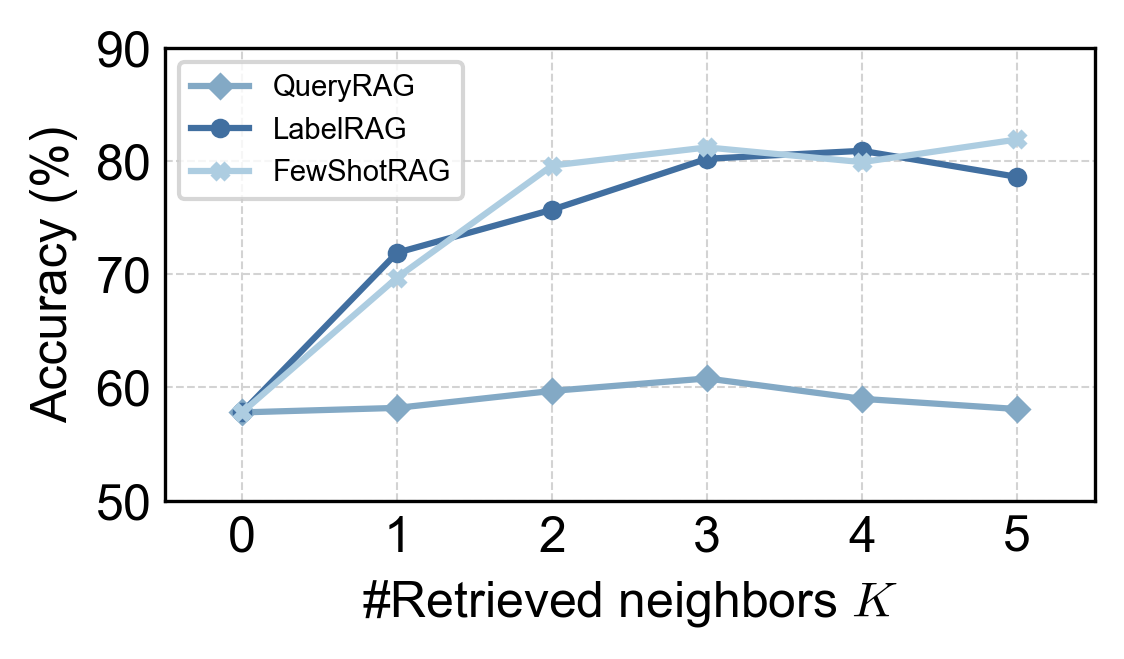}}
  \subfigure[Pubmed]{\includegraphics[width=0.25\linewidth]{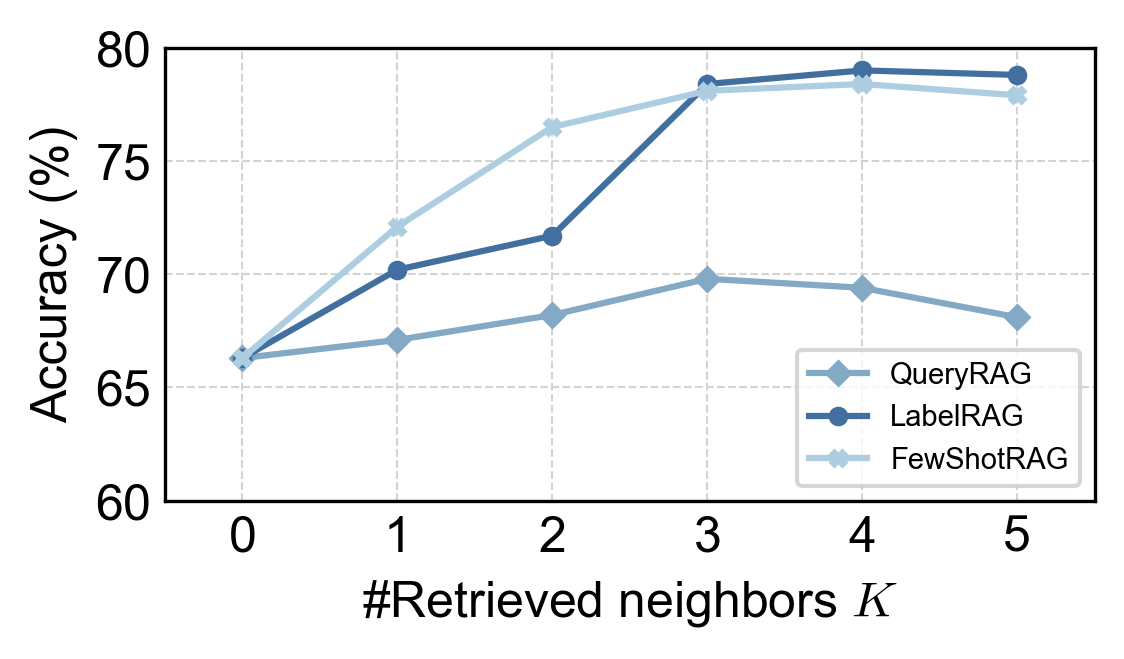}}
  \caption{In-context node classification results on Cora and Pubmed datasets with proposed \qrag, \lrag, and \frag.}
  \label{fig:K}
\end{figure}
\paragraph{Number of retrieved neighbors.}
To investigate the effect of the number of retrieved neighbors $K$ on the performance of our proposed RAG frameworks (\qrag, \lrag, and \frag), we conduct ablation experiments by varying the number of neighbors retrieved for each node. Specifically, we evaluate performance with 0, 1, 2, 3, 4, and 5 retrieved neighbors on the Cora and Pubmed datasets, where  $K=0$  corresponds to the zero-shot setting. We use \textsc{Llama3.1-8B-Instruct} as the LLM model.
The results, shown in Figure~\ref{fig:K}, indicate that increasing the number of retrieved neighbors initially improves performance, as LLMs gain access to more contextual information. However, performance plateaus or even slightly declines beyond a certain point (e.g., more than 3 neighbors), likely due to the introduction of less relevant or redundant information. While retrieving more neighbors can enhance performance by providing richer context, excessively large retrieval sets may dilute the relevance of critical information, highlighting the need for efficient context selection in RAG frameworks.

\section{Additional related work}
We examine recent advancements in applying LLMs to graph data.

\paragraph{LLMs on graphs.}
The integration of graph learning and reasoning with LLMs is a rapidly growing area~\cite{talk_like_graph}. One research direction involves representing graphs as textual data to leverage the reasoning capabilities of pre-trained LLMs. By encoding graph information in a textual format, LLMs can analyze graphs either on-the-fly~\cite{ChenMLJWWWYFLT23,GPT4Graph,WangFHTHT23,instructglm} or through end-to-end fine-tuning~\cite{instructglm,graphllm,GIMLET}. Real-world graphs are often complex, such as text-attributed graphs where nodes contain rich textual information. Another line of research focuses on utilizing the language understanding capabilities of LLMs to extract node features from raw data, which are then fed into GNNs for accurate predictions~\cite{tape,glem}. This approach bridges the strengths of LLMs in processing unstructured text with the structural learning capabilities of GNNs, improving performance on graph-based tasks.
\end{document}